\documentclass{article}

\usepackage{arxiv}

\usepackage[utf8]{inputenc} 
\usepackage[T1]{fontenc}    
\usepackage{url}            
\usepackage{booktabs}       
\usepackage{amsfonts}       
\usepackage{nicefrac}       
\usepackage{microtype}      
\usepackage{lipsum}		
\usepackage{graphicx}
\usepackage{doi}
\usepackage{amssymb,amsmath}

\usepackage{multirow}
\usepackage{caption}
\usepackage{subcaption}
\usepackage{wrapfig}

\title{Dynamic Inter-treatment Information Sharing for Individualized Treatment Effects Estimation}

\author{{\hspace{1mm}Vinod Kumar Chauhan$^1$\thanks{Corresponding author: Vinod Kumar Chauhan (vinod.kumar@eng.ox.ac.uk)\\  accepted to The 27th International Conference on Artificial Intelligence and Statistics (AISTATS) 2024}, Jiandong Zhou$^1$, Ghadeer Ghosheh$^1$, Soheila Molaei$^1$  and David A. Clifton$^{1,2}$} \\
	$^1$Institute of Biomedical Engineering,
	University of Oxford, OX3 7DQ, UK\\
	$^2$Oxford-Suzhou Institute of Advanced Research (OSCAR), Suzhou, China
}

\hypersetup{
}

\begin{document}
\maketitle
\begin{abstract}
    Estimation of individualized treatment effects (ITE) from observational studies is a fundamental problem in causal inference and holds significant importance across domains, including healthcare.
    However, limited observational datasets pose challenges in reliable ITE estimation as data have to be split among treatment groups to train an ITE learner. While information sharing among treatment groups can partially alleviate the problem, there is currently no general framework for end-to-end information sharing in ITE estimation.
    To tackle this problem, we propose a deep learning framework based on `\textit{soft weight sharing}' to train ITE learners, enabling \textit{dynamic end-to-end} information sharing among treatment groups. The proposed framework complements existing ITE learners, and introduces a new class of ITE learners, referred to as \textit{HyperITE}.
    We extend state-of-the-art ITE learners with \textit{HyperITE} versions and evaluate them on IHDP, ACIC-2016, and Twins benchmarks. Our experimental results show that the proposed framework improves ITE estimation error, with increasing effectiveness for smaller datasets.



\end{abstract}


\section{INTRODUCTION}
\label{sec_intro}
While randomized controlled trials (RCT) are the gold standard to ascertain safety and effectiveness of interventions (like policy actions and medical treatments) \cite{pearl2009causality}, they suffer from several limitations, including being expensive, time-consuming, and ethical challenges \cite{bica2021real}. Real-world evidence generated from real-world data, i.e., observational data, can complement existing knowledge as well as address the limitations of RCT, e.g., individualized treatment effects (ITE) estimation for treatment recommendations from electronic health records. ITE is a fundamental problem in \textit{causal inference} that requires predicting counterfactual outcomes \cite{rubin2005causal} for making personalized decisions across various domains. For instance, deciding on the optimal treatment for a patient requires estimating patient outcomes under all treatments \cite{athey2016recursive}, making ITE a crucial aspect of decision-making in healthcare.

Recently, ITE estimation from real-world data has seen great attention from machine learning community, resulting in development of a large number of estimation techniques for personalized treatments \cite{curth2021nonparametric,curth2021inductive,curth2023search,curth2023understanding,chauhan2023adversarial,xue2023assisting,berrevoets2023impute,guo2023estimating,wen2023variational}. ITE estimation problem is different and more complex from standard supervised learning \cite{rubin2005causal,pearl2009causality,wager2018estimation}. This is due to \textit{the fundamental problem of causal inference} of having only factual outcomes and missing counterfactual outcomes for each unit (such as a patient in healthcare) \cite{holland1986statistics}. This complicates shared learning among treatment groups as compared with supervised learning \cite{bica2022transfer}. The primary differences in design choices across the various proposed techniques consider modelling potential outcome functions and addressing the confounding (and resulting selection bias) that exists in the real-world data \cite{curth2021inductive,curth2021nonparametric,kuang2019treatment,kunzel2019metalearners,shi2019adapting,kennedy2022optimal,zhang2021treatment,chauhan2023adversarial,wager2018estimation,nilforoshan2023zero}. To mitigate the effects of selection bias, the existing literature predominantly uses several estimation techniques for the personalized treatment recommendations, including meta-learners (different from meta-learning referring to `learning to learn’) \cite{kunzel2019metalearners,kennedy2022optimal,nie2021quasi,curth2021nonparametric} and representation learning-based learners \cite{johansson2016learning,curth2021nonparametric,shi2019adapting,chauhan2023adversarial,guo2023estimating}.

Nonetheless, for accurate ITE estimation, the training of ITE learners is dependent on the availability of large enough real-world datasets \cite{bica2022transfer}. In domains like healthcare, it can be very hard to gather enough data for rare diseases as only a few patients are available per year \cite{wiens2014study}. Furthermore, at the onset of pandemics, such as COVID-19, scarcity of data limits the ability to understand the efficacy of interventions \cite{qian2020between}. Not only that but as compared with the predictive machine learning setting, the available data have to be split into treatment groups (such as control and treatment in binary treatment setting) further reducing the available data for each treatment group to get personalized treatment recommendations. This is because the existing ITE learners have no general mechanism for end-to-end shared training.
For example, meta-learners train multiple models independently, such as T-Learner trains two independent models for each treatment group in the binary treatment setting \cite{kunzel2019metalearners}, while the representation learning-based techniques learn partly shared representations which are used by treatment-specific layers \cite{shi2019adapting}. Since shared learning among related tasks is known to improve performance, including learning to learn setting \cite{aloui2023transfer,maurer2016benefit} so it becomes essential to develop a general framework for inter-treatment information sharing to improve ITE estimation on limited datasets for reliable personalized treatment recommendations. There is some work on transfer learning with homogeneous feature spaces \cite{kunzel2018transfer} and heterogeneous feature spaces \cite{bica2022transfer}, where information is shared from source domain to target domain in a static way, however, despite some concrete ITE learners, such as \cite{curth2021inductive}, there is no general solution for information sharing among treatment groups.

In this paper, we resolve the problem of lack of a general training framework to share end-to-end information among treatment groups, which is crucial for reliable ITE estimation with limited real-world datasets. We employ soft weight sharing of hypernetworks \cite{ha2017hypernetworks} to propose first deep learning framework for dynamic end-to-end information sharing among the treatment groups to train ITE learners. Hypernetworks, also called as hypernets, are a class of neural networks that generate weights/parameters of another neural network, known as the target network (ITE learner in our case). The training process learns weights of the hypernetwork for data belonging to any treatment group, which then generates weights of different treatment models of an ITE learner, resulting in soft weight sharing among treatment groups (unlike hard-weight sharing achieved through shared layers). The soft weight sharing enables end-to-end sharing of information among treatment groups during each weight update step of training, resulting in dynamic information-sharing among the treatment groups -- helping them to improve ITE estimates (for details refer to Section~\ref{sec_hyper_cate}).
The proposed framework complements existing ITE learners as it provides an alternate way to train, and introduces a new class of ITE learners, called as HyperITE.
We extend state-of-the-art ITE learners from meta-learners as well as representation learning based learners to develop HyperITE versions and evaluate on IHDP, ACIC-2016 and Twins benchmarks that show improved estimates and better performance with smaller dataset sizes.

\section{BACKGROUND}
\label{sec_problem}
The paper addresses the problem of ITE estimation for binary treatments from real-world data using potential outcomes framework \cite{rubin2005causal} for personalized treatment recommendations. Let $D = \lbrace X_i, T_i, Y_i \rbrace_{i=1}^N$ be a sample of $N$ units, such as patients, taken \textit{i.i.d.} from an unknown distribution $\mathbb{P}$. Here, $X_i \in \mathbb{R}^d$ is a $d$-dimensional covariate vector, $T_i \in \lbrace 0, 1 \rbrace$ is a binary treatment variable (with 1 indicating a patient receiving the treatment and 0 not receiving the treatment), and $Y_i$ is patient's outcome that can be binary or real, depending on the ITE estimation problem.

Using the Neyman-Rubin potential outcomes framework \cite{rubin2005causal}, we define $Y_i(1)$ and $Y_i(0)$ as potential outcomes (PO) when patient $i$ receives treatment ($T_i=1$) and when patient does not ($T_i=0$), respectively. However, due to \textit{the fundamental problem of causal inference} \cite{holland1986statistics}, we only observe the factual outcome for the selected treatment, not the counterfactual outcome for the non-selected treatment, i.e., $Y_i = T_i\times Y_i(1) + (1-T_i)\times Y_i(0)$. The ITE is defined as follow, which is also referred to as conditional average treatment effects or heterogeneous treatment effects estimation:
\begin{equation}
	\tau(x) = \mathbb{E}_\mathbb{P} \left[ Y(1) - Y(0) \vert X=x \right],
	\label{eq:cate}
\end{equation}
where $\mathbb{E}_\mathbb{P}$ denotes the expectation with respect to the unknown distribution $\mathbb{P}$.

To estimate ITE from the observational data, we rely on standard assumptions of treatment effect estimation \cite{imbens2015causal}, which are required by the baselines. First, we assume that the \textit{Stable Unit Treatment Value Assumption} (SUTVA) holds, which states that each unit's potential outcome is independent of other units' treatment assignments (no interference) and that all units would receive the same treatment if assigned the same treatment assignment (no hidden variations of treatments). Second, we assume \textit{ignorability}, i.e., the treatment assignment policy is ignorable given the patient's covariate information $X$, i.e.,
\begin{equation}
	\lbrace Y(0), Y(1) \rbrace~\perp~T \vert X,
	\label{eq:ignorability}
\end{equation}
where $\perp$ denotes independence. This assumption is also known as unconfoundedness because it holds if there are no hidden confounders and if additional conditions are satisfied \cite{rosenbaum1983central}. Finally, we assume that the treatment assignment policy is stochastic and that each patient has a certain probability of receiving treatment, i.e.,
\begin{equation}
	0 < \pi(x) < 1, \quad \forall x \in X,
	\label{eq:overlap}
\end{equation}
where $\pi(x)$ is the probability of a patient with covariate $X=x$ receiving treatment $T=1$, called as propensity score. Assumptions (\ref{eq:ignorability}) and (\ref{eq:overlap}) together are called \textit{strong ignorability} assumptions \cite{imbens2009recent}. Under these assumptions, the expected potential outcomes and the ITE are identifiable. Specifically, we have
\begin{equation}
	\label{eq_po}
	\mu_t(x) = \mathbb{E}_\mathbb{P} \left[ Y \vert X=x, T=t \right],
\end{equation}
\begin{equation}
	\label{eq_cate2}
	\begin{split}
		\tau(x) & = \mu_1(x) - \mu_0(x),
	\end{split}
\end{equation}
where $\mu_t(x)$ (or also written as $\mu(x;w_t)$ where $w_t$ are weights) is a PO function of a patient with covariate $X=x$ belonging to treatment group $T=t$.

\section{RELATED WORKS}
\label{sec_literature}
Our work is at the intersection of ITE estimation, information sharing and soft weight sharing. So, we briefly discuss these topics as follows.

\textbf{ITE Learners:} ITE estimation has received great attention from machine learning field due to its ability to handle high dimensional data and complex interactions among the features, as well as due to availability of large observational/real-world datasets, such as electronic health records. This has resulted in the development of myriad ITE learners \cite{johansson2016learning,shalit2017estimating,shi2019adapting,kunzel2019metalearners,curth2021inductive,hassanpour2019counterfactual,tesei2023learning,xue2023assisting,yao2021survey}. Our discussion will primarily focus on `meta-learners' \cite{kunzel2019metalearners}, (which are distinct from meta-learning referring to `learning to learn') and representation learning-based ITE learners \cite{johansson2016learning}. This is due to model-agnostic approach, good theoretical properties and performance of meta-learners, and flexibility, expressiveness and popularity of neural networks \cite{curth2021nonparametric,chauhan2022continuous,chauhan2023brief,chauhan2023hcr}.

\textit{Meta-learners}, originally proposed by \cite{kunzel2019metalearners}, are model-agnostic ITE learners which can work with any machine learning model, and provide a general recipe for estimation. These can be categorized as one-step \textit{plugin learners}, also called as \textit{indirect learners} and two-step \textit{direct learners}. The indirect learners train models for PO functions $\mu_t(x)$ and estimate treatment effect as $\tau(x) = \mu_1(x) - \mu_0(x)$. This category includes S(single)- and T(two)-Learner from \cite{kunzel2019metalearners}. S-Learner augments features with the treatment variable and trains a single model, while T-Learner trains two separate models for each of PO function. On the other hand, two-step direct learners first train models to estimate nuisance parameters $\eta=\{\mu_0, \mu_1, \pi\}$, followed by training models on the pseudo-outcome $Y_{\eta}$ calculated from nuisance parameters $\eta$, to directly estimate treatment effect $\tau$. Different direct learners need different nuisance parameters and have different ways of calculating $Y_{\eta}$ to estimate $\tau$. The well-known direct learners are DR(doubly robust)-Learner \cite{kennedy2022optimal}, RA(regression adjustment)-learner \cite{curth2021nonparametric}, X-Learner \cite{kunzel2019metalearners}, PW(propensity weighting)-learner, and R-Learner \cite{nie2021quasi}.

\textit{Representation learning} based learners use neural networks and multitasking type architecture, and similar to indirect learners, they estimate treatment effect indirectly as $\tau(x) = \mu_1(x) - \mu_0(x)$ by learning $\mu_t(x)$. These learners have shared layers between PO functions $\mu_t(x)$ followed by treatment-specific layers for each $\mu_t(x)$ to get ITE estimates. The joint training of PO functions helps in partial information sharing through the shared layers which learn from the entire data but lack information sharing in PO-specific layers that learn from data corresponding to that treatment group. The most standard architecture and the pioneering work using representation learning is TARNet \cite{johansson2016learning}, which is extended in different ways to balance the covariates to reduce selection bias, such as disentangled representations to permit various types of information sharing between propensity score $\pi(x)$ and PO functions $\mu_t(x)$ \cite{hassanpour2019learning,curth2021nonparametric}. The disentangled representations are based on using orthogonality to learn three or five latent factors for instrument, adjustment and confounder variables, e.g., disentangled representations for counterfactual regression (DRCFR) \cite{hassanpour2019learning} and SNet+ \cite{chauhan2023adversarial}.

\textbf{Information Sharing:} To the best of our knowledge, there is no general mechanism for end-to-end inter-treatment information sharing. Meta-learners \cite{kunzel2019metalearners} train models separately for each of the nuisance parameters $\eta$, including PO functions $\mu_t$ that can access data of corresponding treatment only, and lack any mechanism to share information, except for S-Learner. S-Learner trains a single model and shares information between PO functions $\mu_t$, however, it lacks the flexibility to capture complex treatment effects and is known to perform poorly in high dimensional settings. The representation learning-based learners have partial information sharing as they employ shared layers, however, PO functions have treatment-specific layers which could access data related to the corresponding treatment only \cite{johansson2016learning,curth2021nonparametric}. Transfer learning-based learners allow sharing of information, such as across shared feature spaces \cite{kunzel2018transfer} and heterogeneous feature spaces \cite{bica2022transfer}, however, by definition they consider two related problems or two patient populations for sharing information and do not apply to single problem/population setting. One learner that specifically addresses information sharing is the flexible treatment effect network (FlexTENet) \cite{curth2021inductive} which is based on the inductive biases of having shared structure between PO functions, and have private and shared layers between PO functions. Additionally, disentangled variational auto-encoder (tVAE) \cite{xue2023assisting} is an encoder-decoder based architecture that learns a latent variable where first three elements corresponds to PO functions and propensity score so by architecture have end-to-end information sharing.

\textbf{Soft Weight Sharing:} It is one of the characteristics of hypernetworks arising when there are multiple target networks, like PO functions in our case. \textit{Hypernetworks}, or hypernets in short, also referred to as hypermodels, meta-networks, or naturally meta-models, is the term coined by Ha et al. \cite{ha2017hypernetworks}, although, similar ideas were discussed earlier, e.g., \textit{fast weights} in \cite{schmidhuber1992learning}. These are a class of neural networks which generate weights of a target neural network. They are composed of two networks: a primary network (hypernet) and a target network (ITE learner in our case). The hypernet takes an identity/embedding of PO functions of an ITE learner and learns to generate weights of ITE learner. Both the networks are trained end-to-end, however, only the hypernet has trainable weights. Since hypernet generates the weights for all potential outcome functions so act as soft weight sharing (unlike hard weight sharing where layers are shared) among PO functions and facilitates information sharing among them.
Hypernetworks have recently emerged as a promising approach for improving the expressiveness and flexibility of deep learning models, and are successful in providing state-of-the-art results across different problem settings, including in multitasking \cite{tay2021hypergrid}, federated learning \cite{litany2022federated}, continual learning \cite{Oswald2020Continual}, ensemble learning \cite{kristiadi2019predictive}, neural architecture search \cite{peng2020cream}, camera pose localization \cite{ferens2023hyperpose}, weight pruning \cite{liu2019metapruning}, hyperparameter optimization \cite{lorraine2018stochastic}, Bayesian neural networks \cite{deutsch2019generative}, generative models \cite{deutsch2019generative}, knowledge distillation \cite{wu2023hyperinr}, neural style transfer \cite{Ruta2023}, quantum computing \cite{carrasquilla2023quantum}, density estimation \cite{Hofer2023}, the Pareto-front learning \cite{tran2023framework}, and adversarial defence \cite{sun2017hypernetworks}. However, to the best of our knowledge, soft weight sharing and hypernets are not used in treatment effect estimation.

Thus, based on our brief review of the related work, we conclude that information sharing among treatment groups for ITE estimation remains an open challenge. While there are studies on transfer learning in both heterogeneous and homogeneous feature spaces, the field lacks a general framework for end-to-end information sharing. We propose the first general framework for dynamic end-to-end inter-treatment information sharing for ITE estimation, which similar to transfer learning is crucial for limited observational datasets.

\section{DYNAMIC INTER-TREATMENT INFORMATION SHARING}
\label{sec_hyper_cate}
Motivated by the proven benefits of information sharing among related tasks, such as transfer learning \cite{aloui2023transfer} and learning to learn settings \cite{maurer2016benefit}, and the lack of a general framework for information sharing among treatment groups in treatment effects estimation, we introduce a deep learning framework based on soft weight sharing\footnote{This is soft weight sharing because unlike the hard-weight sharing where PO functions directly share weights, here weights are shared indirectly to generate weights for PO functions.}. This framework enables dynamic information sharing and enhances the reliability of ITE estimation, especially in scenarios with limited observational data. The proposed framework provides an alternative way of training existing ITE learners, resulting in a new class of learners called as HyperITE.

Suppose $f_{ITE}(\theta)$ is an ITE learner with weights $\theta$ and $(\mu_0(x), \mu_1(x))$ are potential outcome (PO) functions corresponding to treatment $T=0$ and $T=1$, respectively. $\mathcal{H}$ is a hypernet with weights $\psi$ that generates weights $\theta$ of $f_{ITE}$, i.e., $\theta = \mathcal{H}(\cdot;\psi)$ by taking an embedding vector $e$ as input corresponding to PO functions. In standard training of an ITE learner, the training process optimizes its weights $\theta$ directly from data, while in HyperITE, i.e., an ITE learner trained using hypernets, only hypernet weights $\psi$ and not ITE learner weights $\theta$ are trainable. So, optimization problems for ITE and HyperITE are as given below.
\begin{equation}
	\begin{array}{l}
		\text{ITE:} \quad \quad \quad \quad \quad \min_{\theta}\quad f_{ITE}(\theta), \\ 
		\text{HyperITE:} \quad \min_{\psi}\quad f_{ITE}(\theta=\mathcal{H}(\cdot;\psi)).
	\end{array}
\end{equation}


For HyperITE learners, data belonging to all treatment groups lead to gradients flow to update the hypernet weights in each training step -- resulting in dynamic inter-treatment information sharing during the training of an ITE learner, as explained in following subsections.
We illustrate HyperITE versions of following two categories of ITE learners: meta-learners \cite{kunzel2019metalearners,curth2021nonparametric} which have no information sharing, and representation learning-based learners \cite{shalit2017estimating,chauhan2023adversarial} that have partial information sharing, and the framework enables them to share end-to-end information between PO functions.

\subsection{HyperITE for Meta-learners}
\label{subsec_method_meta_learners}
\begin{figure*}[htb!]
	\centering
	\includegraphics[width=0.95\textwidth]{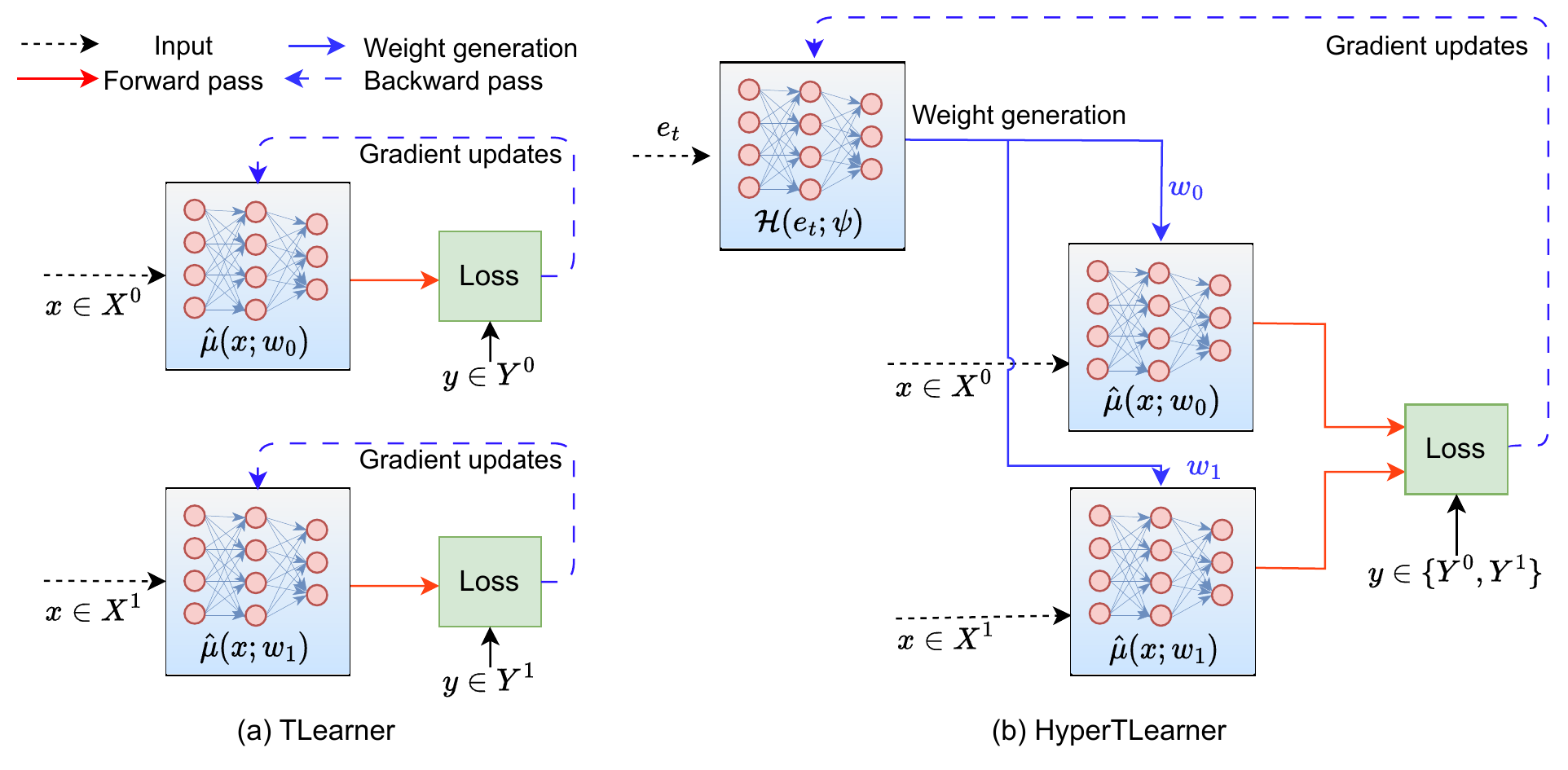}
	\caption{An overview of the architectures and gradient flows for T-Learner and HyperTLearner, where $\hat{\mu}(x;w_0) \text{ and } \hat{\mu}(x;w_1)$ have exactly the same architecture in both but different training process.}
	\label{fig_HyperTLearner}
\end{figure*}

To illustrate application of the proposed framework to train existing meta-learners, we take an example of T-Learner due to ease of application and its lack of any information sharing as it trains potential outcome functions $\mu_0(x)$ and $\mu_1(x)$ independently.
Fig.~\ref{fig_HyperTLearner} presents a comparative view of T-Learner and HyperTLearner, i.e., T-Learner trained with the proposed framework, with architectures and flow of gradients. As shown in Fig.~\ref{fig_HyperTLearner}(a), T-Learner splits dataset $\{X_i,T_i,Y_i\}$ according to treatment into two groups as $\{X_i^0,T_i=0,Y_i^0\}$ and $\{X_i^1,T_i=1,Y_i^1\}$ for estimating $\mu_0(x), \text{ and } \mu_1(x)$, respectively. Each PO function has its own gradient flow and independent weights $w_t$ updates, lacking any information sharing. On the other hand, in HyperTLearner in Fig.~\ref{fig_HyperTLearner}(b), the exact same T-Learner is trained using soft weight sharing.  Hypernet $\mathcal{H}$ takes an embedding $e_t$ and generates weights $w_t$ for the corresponding PO \textit{functional} $\mu(x;w_t)$, i.e., $w_t = \mathcal{H}(e_t;\psi)$. Next, a forward pass through any of $\mu(x;w_0)$ or $\mu(x;w_1)$ results in gradients flowing back to hypernet $\mathcal{H}$ to update weights $\psi$ in each step. Thus, soft weight sharing between $\mu_0(x)$ and $\mu_1(x)$ through hypernet weights $\psi$ facilitates continuous inter-treatment information sharing as $\psi$ are updated using data belonging to both treatment groups, unlike the T-Learner where $\mu_0(x)$ and $\mu_1(x)$ could use data corresponding to that treatment group only. Thus, the training process optimizes hypernet weights $\psi$ so that it generates T-Learner's weights correctly to reduce prediction errors, and resulting into dynamic end-to-end information sharing between treatment groups.

Using similar approach, we can extend the proposed framework to other meta-learners. For example, DR-Learner \cite{kennedy2022optimal} is a two-step learner where it estimates nuisance parameters $\eta = (\mu_0(x), \mu_1(x), \pi(x))$ in first step which are then used to calculate pseudo-outcomes $Y_{\eta}$. In second step, $Y_{\eta}$ is regressed on the context $X$ to directly estimate treatment effect $\tau(x)$. To develop HyperDRLearner, we use hypernet in the first step to generate three networks corresponding to $(\mu_0(x), \mu_1(x), \pi(x))$, instead of two in HyperTLearner, while the second step remains similar to DR-Learner as it trains a single network.

\subsection{HyperITE for Representation Learning-based Learners}
\label{subsec_method_representationsNN}
\begin{figure*}[htb!]
	\centering
	\includegraphics[width=0.95\textwidth]{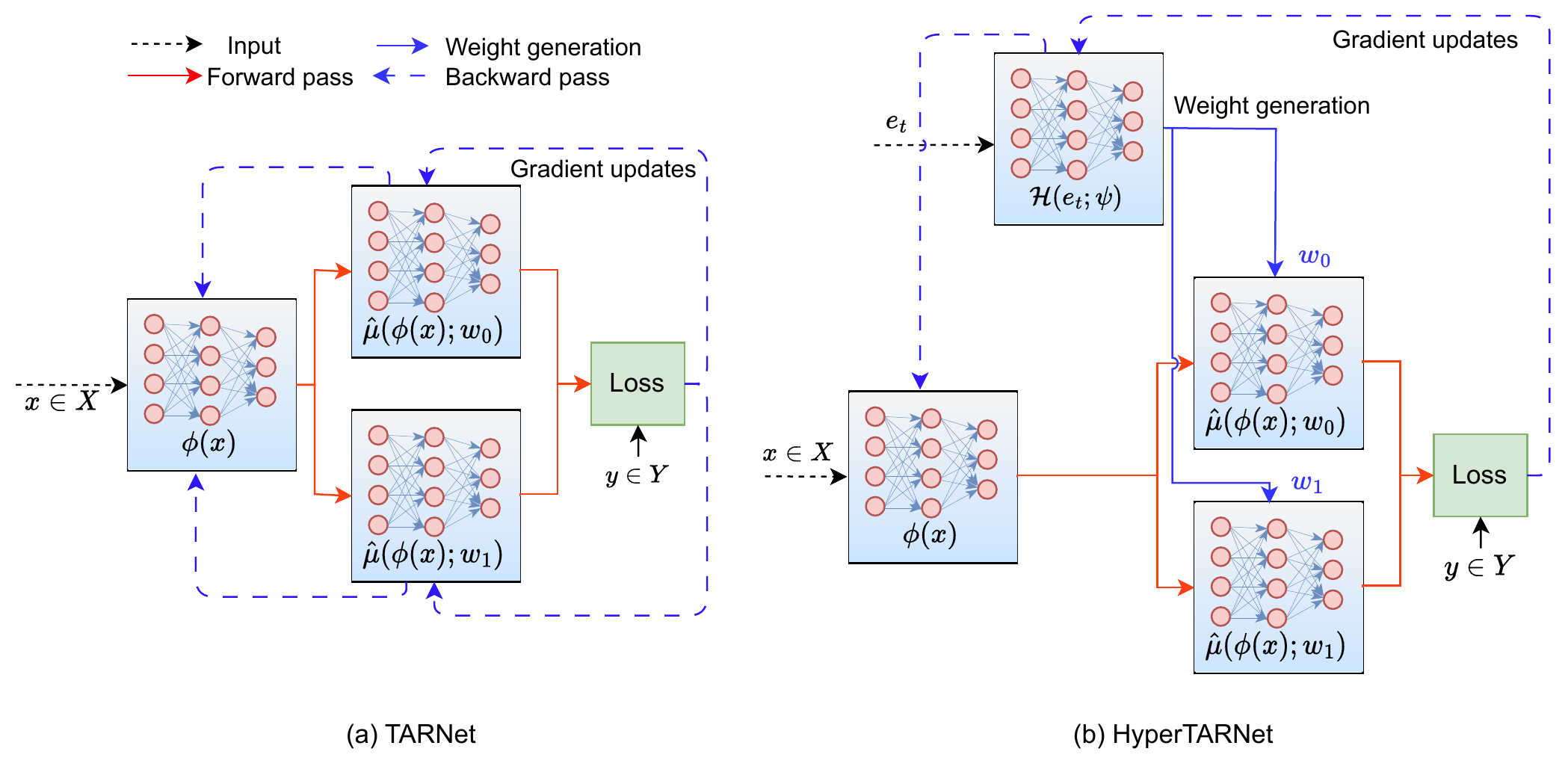}
	\caption{Architectures and gradient flows for TARNet and HyperTARNet, where $\hat{\mu}(x;w_0) \text{ and } \hat{\mu}(x;w_1)$ have exactly the same architecture in both but different training process. HyperTARNet is employed to train and share information between PO heads, while $\phi(x)$ is learnt similarly to TARNet.}    
	\label{fig_HyperTARNet}
\end{figure*}

Shalit et al. \cite{shalit2017estimating} pioneered the idea of representation learning to address the selection bias in treatment effect estimation where they presented a simple architecture having shared layers (hard weight sharing) between PO functions, to learn shared representation $\phi(x):\mathbb{R}^d\rightarrow\mathbb{R}^{d_r}$ from $d$-dimensional input to $d_r$-dimensional representation, but treatment-specific layers for $\mu_0(\phi(x)), \mu_1(\phi(x))$, resulting in TARNet architecture. We extend the proposed framework to TARNet due to its popularity and having simple architecture among representation learning based learners, and illustrate differences between TARNet and HyperTARNet in Fig.~\ref{fig_HyperTARNet}.

As shown in the figure, during forward pass, TARNet learns a shared representation $\phi(x)$ for each input $x$, which then passes through treatment specific layer to get either $\mu_0(\phi(x)) \text{ or } \mu_1(\phi(x))$. This outcome along with factual outcome $y$ is used for calculating loss. During the backward pass, gradients flow to $\phi(x)$ through the treatment specific layer $\mu_0(\phi(x)) \text{ or } \mu_1(\phi(x))$, and weights are updated for $\phi(x)$ and $\mu_0(\phi(x)) \text{ or } \mu_1(\phi(x))$ depending on $x \in X^0$ or $x \in X^1$. So, TARNet has partial information sharing due to hard weight sharing through shared layers. On the other hand, in HyperTARNet, during the forward pass, hypernet $\mathcal{H}$ takes an embedding $e_t$ of treatment specific head and generates weights $w_t$. Moreover, similar to TARNet, HyperTARNet takes an input $x$ to learn a representation $\phi(x)$ which is then passed through treatment specific head, i.e., $\mu_0(\phi(x)) \text{ or } \mu_1(\phi(x))$. $\mu_t(\phi(x))$ for treatment $t$ and factual outcome $y$ are used to calculate loss which then backpropagates gradients to hypernet $\mathcal{H}$ for updating the weights $\psi$. Data corresponding to both treatments groups results in updating of $\psi$ in each training step, and hence dynamically sharing end-to-end information between $\mu_0(x), \text{ and } \mu_1(x)$, unlike TARNet which has partial information sharing.

There are several extensions of TARNet for handling the selection-bias, such as most recently SNet+ \cite{chauhan2023adversarial}. Similar to TARNet, the proposed HyperITE framework can be easily extended to train these representation learning based learners. Please refer to supplementary material for one more example of application of the framework to SNet+ which is relatively a complex learner, as well as details of remaining HyperITE versions.



\section{EVALUATION}
\label{sec_results}
In this section, we present experimental settings and results to prove efficacy of the proposed framework of dynamic end-to-end information sharing between $\mu_0(x), \text{ and } \mu_1(x)$.

\subsection{Experimental Settings}
\label{subsec_exp_settings}
\textbf{Benchmarks:} We consider two semi-synthetic and one real-world benchmarks: Infant Health and Development Program (\textit{IHDP}) dataset, 2016 Atlantic Causal Inference Conference (\textit{ACIC-2016}) Competition dataset and \textit{Twins} dataset, respectively. IHDP dataset \cite{hill2011bayesian,shalit2017estimating} for causal inference is a binary treatment problem which has real covariates, however, outcomes are simulated. It is a small dataset with 139 treated and 608 untreated units (a total of 747 units) and 25 covariates. The child care/specialist home visits are regarded as treatments, the future cognitive test scores of the children are outcomes, while the covariates are features measured from both mother and child. On the other hand, ACIC-2016 dataset \cite{dorie2019automated}, initially made available as part of the competition and obtained from the Collaborative Perinatal Project, comprises data for 2200 patients. This dataset contains a total of 55 covariates, including both continuous and categorical variables. We utilized the preprocessing pipeline from the \textit{CATENets} package\footnote{https://github.com/AliciaCurth/CATENets}, which results in 4000 and 802 points in train and test sets with 55 features. Twins \cite{almond2005costs} is a real-world dataset which collects 11,400 twin births in the US between 1989 to 1991. It consists of 39 covariates related to the parents, pregnancy, and birth. Being heavier at birth is treatment and one-year mortality is considered outcome, i.e., this is a binary treatment problem with binary outcomes. We followed \textit{CATENets} preprocessing and split the data as 50:50 for train and test sets. Moreover, we sample the heavier twin with a probability of 0.1 to have imbalanced data because ITE learners have less variability in the balanced cases.

\begin{table*}[htb!]
	\small
	\centering
	\caption{Comparative study of ITE learners against HyperITE learners on IHDP, ACIC-2016 and Twins benchmarks using PEHE (lower is better) as a performance metric. Each experiment is averaged over multiple runs and values in parenthesis refer to one standard error.}
	\label{tab_results}
	\begin{tabular}{lrrrrrr}\\ \hline
		\multicolumn{1}{c}{\multirow{2}{*}{\textbf{Learner}}} &
		\multicolumn{2}{c}{\textbf{IHDP}} &
		\multicolumn{2}{c}{\textbf{ACIC-2016}} &
		\multicolumn{2}{c}{\textbf{Twins}} \\
		\multicolumn{1}{c}{} &
		\multicolumn{1}{c}{\textbf{PEHE-in}} &
		\multicolumn{1}{c}{\textbf{PEHE-out}} &
		\multicolumn{1}{c}{\textbf{PEHE-in}} &
		\multicolumn{1}{c}{\textbf{PEHE-out}} &
		\multicolumn{1}{c}{\textbf{PEHE-in}} &
		\multicolumn{1}{c}{\textbf{PEHE-out}} \\ \hline
		SLearner       & 1.02 (0.001) & 0.97 (0.001) &  3.76 (0.005) & 4.00 (0.010) & 0.318	(0.001) &	0.331	(0.001)\\
		HyperSLearner  & \textbf{0.99 (0.001)} & \textbf{0.95 (0.001)} &  \textbf{3.75 (0.002)} & \textbf{3.96 (0.005)} & \textbf{0.309	(0.001)} &	\textbf{0.324	(0.001)}\\ \hline
		TLearner       & 1.46 (0.002) & 1.32 (0.003) &  5.37 (0.019) &	5.14	(0.045) & 0.398	(0.001) &	0.410	(0.001) \\
		HyperTLearner  & \textbf{1.21 (0.002)} & \textbf{1.12 (0.001)} &  \textbf{4.40 (0.053)} & \textbf{4.51 (0.031)} & \textbf{0.332	(0.003)}	& \textbf{0.345	(0.003)} \\ \hline
		DRLearner      & 1.18 (0.002) & 1.12 (0.002) & \textbf{4.28 (0.061)} & 4.81 (0.041) & \textbf{0.307 (0.001)} &	\textbf{0.323 (0.001)}\\
		HyperDRLearner & \textbf{1.16 (0.002)} & \textbf{1.09 (0.002)} & 4.42 (0.048) & \textbf{4.48 (0.071)} & 0.308	(0.001) &	\textbf{0.323	(0.001)}\\ \hline
		RALearner      & 1.41 (0.001) & 1.25 (0.002) &  5.22 (0.025) & 4.66 (0.026) & 0.373	(0.001)	& 0.384	(0.001) \\
		HyperRALearner & \textbf{1.17 (0.002)} & \textbf{1.06 (0.002)} &  \textbf{4.45 (0.054)} & \textbf{4.33 (0.028)} & \textbf{0.314	(0.001)}	& \textbf{0.329	(0.001)} \\ \hline
		TARNet         & 1.36 (0.001) & 1.24 (0.002) &  4.97 (0.045) & 4.79	(0.040) & 0.355	(0.001) &	0.366	(0.001) \\
		HyperTARNet    & \textbf{1.15 (0.002)} & \textbf{1.08 (0.002)} &  \textbf{4.06 (0.040)} & \textbf{4.13 (0.038)} & \textbf{0.324	(0.003)} &	\textbf{0.338	(0.002)} \\ \hline
		MitNet       & 1.36 (0.004) & 1.25 (0.004) &  4.93	(0.043) &	4.76	(0.043) & 0.309	(0.012) &	0.324	(0.012) \\
		HyperMitNet  & \textbf{1.18 (0.009)} & \textbf{1.12	(0.005)} &  \textbf{4.03 (0.041)} & \textbf{4.09 (0.041)} & \textbf{0.308	(0.011)} &	\textbf{0.323	(0.012)} \\ \hline
		SNet+          & 1.35 (0.001) & 1.22 (0.001) &  5.48 (0.013) & 5.03 (0.013) & 0.339	(0.003) &	0.353	(0.003)\\
		HyperSNet+     & \textbf{1.29 (0.001)} & \textbf{1.17 (0.001)} &  \textbf{5.09 (0.020)} &	\textbf{4.85 (0.015)}  & \textbf{0.310	(0.001)} &	\textbf{0.325	(0.001)}\\ \hline
		FlexTENet      & 1.32 (0.011) & 1.19 (0.011) &  4.86 (0.042) &	4.70 (0.043) & 0.313	(0.000)	& 0.328	(0.000)  \\
		\hline
	\end{tabular}
\end{table*}

\textbf{Baselines:} Following \cite{bica2022transfer}, we have used DR-Learner \cite{kennedy2022optimal}, S-Learner \cite{kunzel2019metalearners}, T-Learner \cite{kunzel2019metalearners} and TARNet \cite{johansson2016learning} as our baselines. Additionally, we have also considered FlexTENet \cite{curth2021inductive} as a baseline because it proposes a flexible architecture for information sharing, RA-Learner \cite{curth2021nonparametric} due to its suitability to the framework and importance \cite{nilforoshan2023zero}, and recent representation learning-based learners as SNet+ \cite{chauhan2023adversarial} and MitNet \cite{guo2023estimating}. We present more results with tVAE \cite{xue2023assisting} in supplementary material. We have compared the baselines with their HyperITE variants which enable baselines to have dynamic end-to-end inter-treatment information sharing. For a fair comparison between ITE and HyperITE, both have exactly the same architecture and training loop. Moreover, we follow \cite{crabbe2022benchmarking} to set basic hyperparameters. All learners have two hidden layers of 100 neurons each, and for the sake of simplicity, hypernets also have two hidden layers of 100 neurons each (for details please refer to supplementary material Section 2. Implementation Details). We utilize Precision in the Estimation of Heterogeneous Effects (PEHE) \cite{hill2011bayesian} $\sqrt{N^{-1} \sum_{i=1}^N \left(\left(\hat{\mu}_1^i - \hat{\mu}_0^i\right) - \left(\mu_1^i - \mu_0^i\right)\right)^2}$ as a metric to evaluate the performance of the ITE learners. This is because semi-synthetic and real datasets contain both factual and counterfactual outcomes, and treatment effects can be calculated.
All the experiments are implemented in Python using PyTorch as a deep learning framework. The experiments are executed on an Ubuntu machine (64GB RAM, one NVIDIA GeForce GPU 8GB) and each experiment is averaged using ten runs, except for IHDP where we averaged over 100 runs due to smaller size, and reports one standard error.

\subsection{Results}
\label{subsec_results}
Table~\ref{tab_results} presents a comparative study of HyperITE against the baselines on IHDP, ACIC-2016 and Twins benchmarks using PEHE as a performance metric for ITE estimation. Here, FlexTENet does not have a HyperITE version because it has a flexible architecture that provides end-to-end information sharing. PEHE-in refers to PEHE (in distribution) on train data and PEHE-out (out distribution) refers to PEHE on a held-out test data.
On IHDP dataset, all HyperITE perform better than the corresponding baselines as well as FlexTENet for both PEHE-in and -out metrics.
For ACIC-2016 and Twins benchmarks, HyperITE performs better than the baselines, except for HyperDRLearner, where HyperDRLearner performs better on the out-distribution of data but not on in-distribution. One potential reason for the poor performance of HyperDRLearner, which is a two-step learner, is the regularization of the second-step models by the errors in the first step, especially by the propensity score $\hat{\pi}$ whose extreme probabilities can even destabilize the training.
HyperSLearner is the best performer on IHDP and ACIC-2016, and close to the best performance on Twins dataset, however, the performance improvement is very small. This is because S-Learner trains a single learner. DR-Learner performs the best on Twins dataset and closely followed by HyperDRLearner and HyperMitNet.
The performance of ITE learners on PEHE-in and -out show similar patterns.
In general, HyperTLearner and HyperRALearner show the highest improvements over the corresponding ITE learner across all the datasets. This is because the corresponding baselines do not have any mechanism to share information between $\mu_0(x)$ and $\mu_1(x)$ while HyperITE variants provide end-to-end information sharing.
On comparing representation learning and meta-learner based HyperITE learners, we observe that representation learning-based HyperITE show clear improvements over corresponding baselines, however, for meta-learners one side there are large gains for HyperTLearner (direct learner) and HyperRALearner (in-direct learner) but on the other side HyperDRLearner does not show clear pattern.

\begin{figure}[htb!]
	\centering
	\begin{subfigure}[b]{0.49\textwidth}
		\centering
		\includegraphics[width=\textwidth]{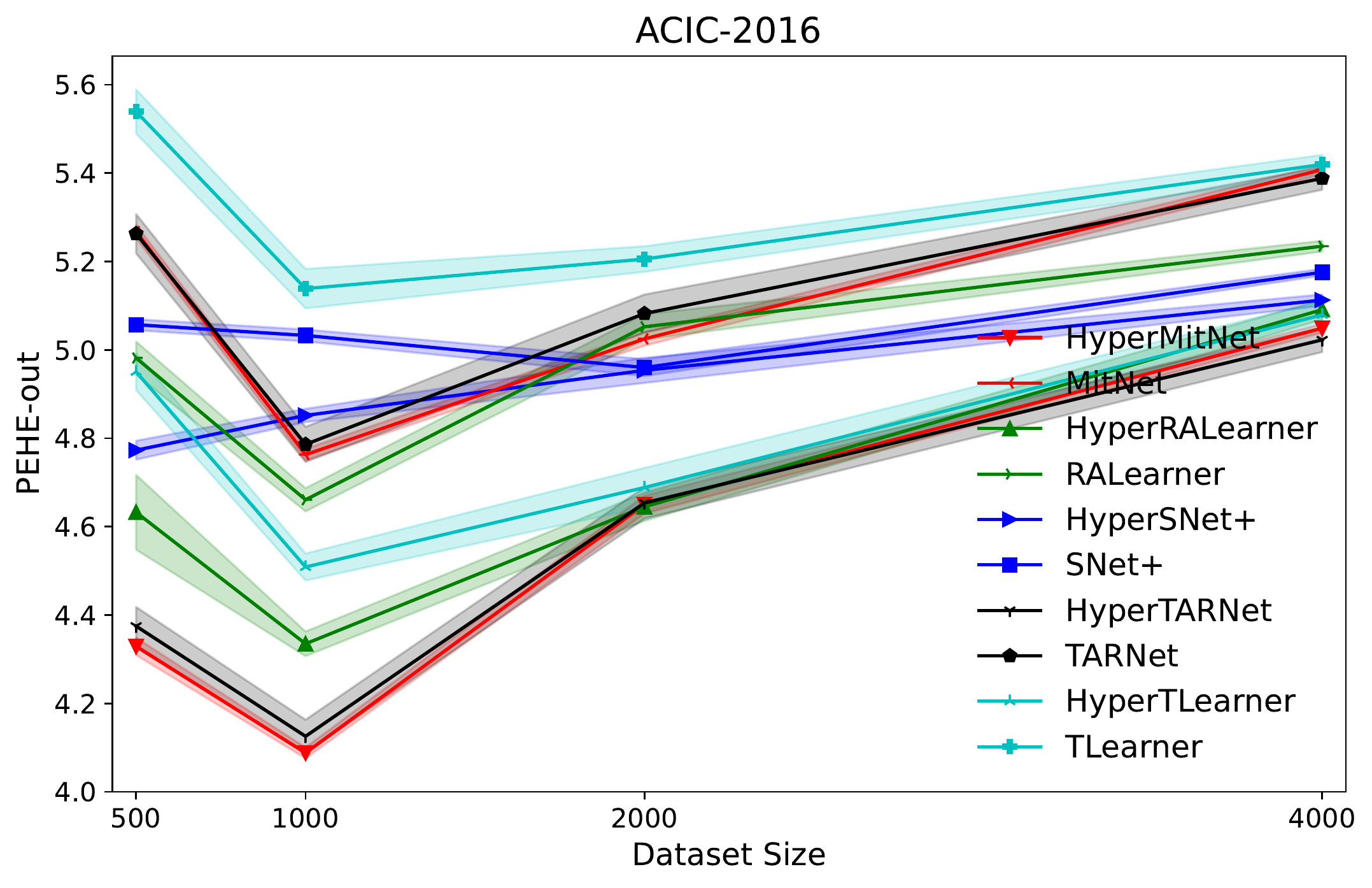}
	\end{subfigure}
	\begin{subfigure}[b]{0.49\textwidth}
		\centering
		\includegraphics[width=\textwidth]{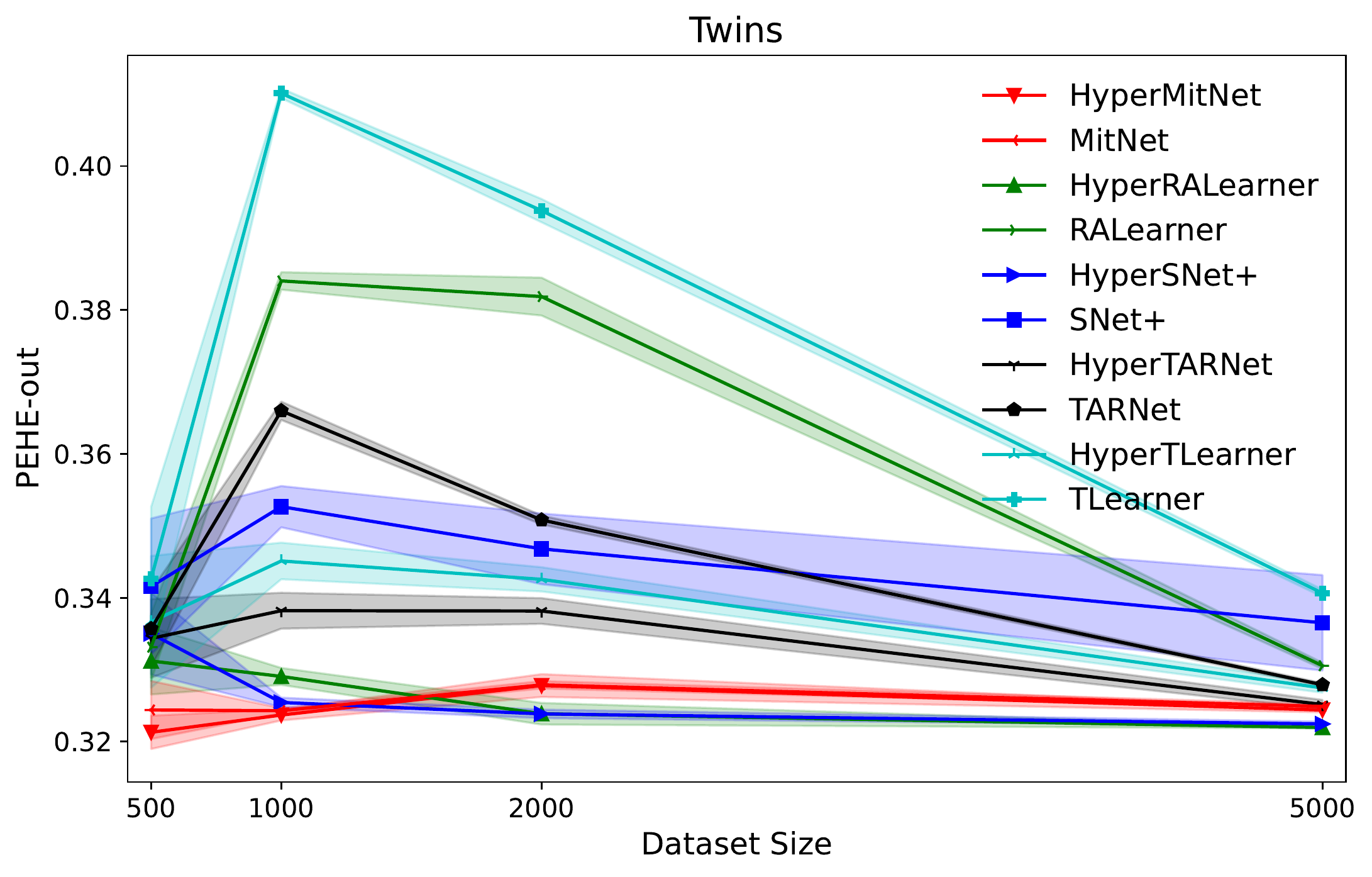}
	\end{subfigure}
	\caption{Effect of dataset size on the performance of selected learners using ACIC-2016 and Twins datasets (shaded region shows one standard error).}
	\label{fig_scale}
\end{figure}

\textbf{Effect of Limited Observational Datasets:} In Table~\ref{tab_results}, we studied the performance of HyperITE with one thousand data points of ACIC-2016 and Twins datasets, to study effectiveness of HyperITE for limited real-world data settings. Fig.~\ref{fig_scale} depicts effect of dataset size on HyperITE performance using ACIC-2016 and Twins datasets (we can't do similar experiments for IHDP due to its small size). For the sake of keeping figures tidy, we keep only those HyperITEs which showed good results on three datasets in Table~\ref{tab_results}, and we study dataset size effect on those to find if they show consistent performance (please refer to supplementary material for the rest of the learners). Moreover, the performance curves are almost similar for PEHE-in and -out so we present results using PEHE-out (please refer to supplementary material for results with PEHE-in).

From Fig.~\ref{fig_scale} for both the datasets, we find that all HyperITEs show consistent performance over the corresponding baselines. Moreover, overall the performance curves of learners diverge towards the left side (except for Twins at 500 data points), i.e., with decreasing dataset size the performance gap between HyperITE and baselines increases. This proves effectiveness of HyperITEs for limited real-world data setting for estimation of ITE for personalized treatment recommendations. Please refer to supplementary material for additional experiments on effect of embedding size on performance of HyperITE, effect of different weight generation strategies in HyperITE, and convergence of ITE against HyperITE.

\section{CONCLUSION}
\label{sec_conclusion}
In this paper, we solved an important issue of information sharing among treatment groups in ITE estimation, which is crucial while studying interventional effects on limited observational datasets. We proposed \textit{HyperITE}, a new class of ITE learners, which complements the existing ITE learners and provides a general framework to train neural network based ITE learners, such as meta-learners and representation learning-based learners. The proposed framework provides dynamic end-to-end information sharing across treatment groups, unlike the existing learners which lack such a mechanism. HyperITE utilizes \textit{soft weight sharing} for dynamic information sharing across treatment groups. We validated the proposed framework with IHDP, ACIC-2016 and Twins benchmarks, and empirically proved their effectiveness, especially for small datasets.

One limitation of HyperITE is the convergence issue of hypernetworks \cite{chang2020Principled} for generating weights of potential outcome (PO) functions. Similar to multi-tasking \cite{liu2021conflict}, hypernets can suffer from conflicting gradients among the treatment groups, unlike standard ITE learners, such as T-Learner which trains PO functions separately and do not face such issues. A proper mechanism for handling the convergence of hypernets is still an open issue, however, adaptive optimizers such as Adam can address the issue to some extent \cite{chang2020Principled}. Another limitation of the proposed mechanism is that it adds extra hyperparameters which have to be tuned for each dataset, although these can be tuned along with other machine-learning hyperparameters on the validation dataset.
Finally, we acknowledge that causal assumptions may not always hold in real-world scenarios, potentially leading to biased results. Therefore, it is crucial to validate causal inference models rigorously. These models should be used for decision support only in collaboration with clinicians due to the possible severe consequences in healthcare settings.

\section*{Acknowledgments}
This work was supported in part by the National Institute for Health Research (NIHR) Oxford Biomedical Research Centre (BRC) and in part by InnoHK Project Programme 3.2: Human Intelligence and AI Integration (HIAI) for the Prediction and Intervention of CVDs: Warning System at Hong Kong Centre for Cerebro-cardiovascular Health Engineering (COCHE).
DAC was supported by an NIHR Research Professorship, an RAEng Research Chair, the InnoHK Hong Kong Centre for Cerebro-cardiovascular Health Engineering (COCHE), the NIHR Oxford Biomedical Research Centre (BRC), and the Pandemic Sciences Institute at the University of Oxford. 
The views expressed are those of the authors and not necessarily those of the NHS, the NIHR, the Department of Health, the InnoHK – ITC, or the University of Oxford.

\subsection*{Code Availability}
The code is is available at \textit{\url{https://github.com/jmdvinodjmd/HyperITE}}.

\subsection*{Contributions}
\textbf{V.K.C.}: Conceptualization, Methodology, Software and Writing- Original draft preparation. \textbf{J.Z.}: Writing - Review \& Editing. \textbf{G.G.}: Writing - Review \& Editing. \textbf{S.M.}: Writing - Review \& Editing. \textbf{D.A.C}: Writing - Review \& Editing, Supervision and Funding acquisition.


\clearpage
\newpage
\appendix

\section{AN ADDITIONAL EXAMPLE of ITE vs HYPERITE}
\label{app_additional_example}

\begin{figure}[htb!]
		\centering
		\includegraphics[width=\linewidth]{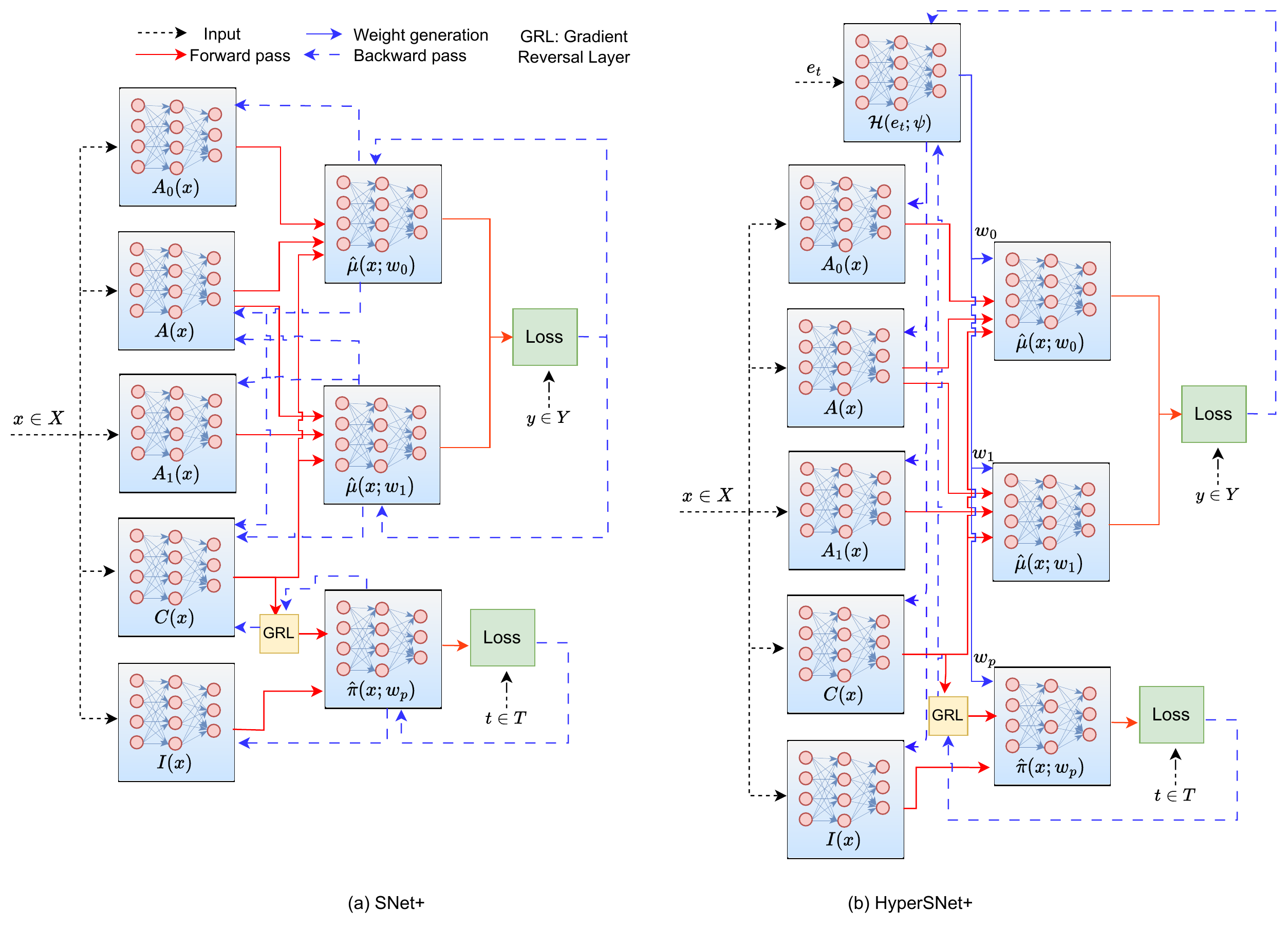}
		\caption{An overview of the architectures and gradient flows for SNet+ and HyperSNet+, where $\hat{\mu}(x;w_0), \hat{\mu}(x;w_1) \text{ and } \hat{\pi}(x;w_p)$ have exactly the same architecture in both but different training process.}
		\label{fig_snet}
	\end{figure}

Fig.~\ref{fig_snet} provides another example of application of the proposed framework to SNet+ \cite{chauhan2023adversarial}, and compares the architecture and gradient flows of SNet+ against HyperSNet+.
SNet+ is presented in subfigure (a), where weights $w_0$ and $w_1$ of potential outcome (PO) function estimators $\hat{\mu}_0$ and $\hat{\mu}_1$ and $w_p$ for propensity score estimator $\hat{\pi}$ are learnable, and are trained through data points corresponding to the treatment group, i.e., $\hat{\mu}_0$ learns only from control group $T=0$ and $\hat{\mu}_1$ learns only from treatment group $T=1$. SNet+ has partial information sharing through latent factors, however, the PO functions have different gradient flows so SNet+ lacks end-to-end information sharing. On the other hand, for HyperSNet+ in subfigure (b), weights $w_0, w_1$ and $w_p$ are not learnable as they are generated by hypernet. The data corresponding to any treatment group leads to updating weights of hypernet parameters during each learning step and this soft-weight sharing between PO functions enables dynamic end-to-end information sharing between treatment groups.

\section{IMPLEMENTATION DETAILS}
\label{app_implementation}
We follow \cite{crabbe2022benchmarking} to set the basic hyperparameters. All learners have two hidden layers of 100 neurons each, and for the sake of simplicity, the hypernet also has two hidden layers of 100 neurons each. Neither can one ITE learner perform well for all the problems \cite{machlanski2023hyperparameter} nor do we intend to compare them. Moreover, it is difficult to set similar architecture for all the learners. But this does not create any bias in our results because our comparison is between ITE and HyperITE, and for both cases the learner always has the exact same architecture and training loop but with a different training approach. So, S-Learner and T-Learner have two layers with 100 neurons each, TARNet has one common layer of 100 neurons and each treatment specific head has one hidden layer with 100 neurons. MitNet has architecture similar to TARNet, and for discrete treatments, it uses adversarial training. For tVAE, we use official implementation\footnote{https://github.com/xuebing1234/tvae} with two hidden layers of 100 neurons each for encoder as well as decoder. Similarly, for all direct learners, each model in the first and second stage have two layers with 100 neurons each. For SNet+, we use a hidden layer with 50 neurons in each of the five multi-layer perceptrons (MLPs) to learn latent factors for confounders, instrument and adjustment variables. The treatment-specific layers have one hidden layer with 100 neurons each. For FlexTENet, private and shared layers have two hidden layers with 50 neurons each. All ITE learners use ReLU as an activation function in the hidden layers of ITE learners as well as hypernet. We have used Adam as an optimizer \cite{kingma2014adam} with a learning rate of 0.0001, weight-decay of 0.0001, early stopping with the patience of 50 and mini-batch size of 1024. We have used a validation set as 30\% of the train set. For HyperITE, we fix dropout of 0.05 in the hypernetwork -- a small value is used to avoid disruptive changes in the output weights. Moreover, we use the spectral-norm layer \cite{Miyato2018spectral} in the hypernet network to stabilize the training. All experiments use generate once hypernet that generates all the weights of an ITE learner at once. The gradient reversal parameter in adversarially trained ITE learners, like SNet+, is gradually increased during training by setting $\gamma$ to 300. For FlexTENet, we set $\lambda_1, \lambda_2$ and orthogonality regularisation factor to 0.0001, 0.01 and 0.1, respectively. We use embeddings of size 1, 8, 8, 8, 8, 16 and 16 for HyperSLearner, HyperTLearner, HyperTARNet, HyperSNet+, HyperRALearner, HyperDRLearner and HyperXLearner, respectively, and are implemented inside the hypernet, i.e., hypernet takes the identity of the PO function and learns an embedding. It is an implementation choice as the embeddings can be put outside the hypernet. In that case, it would take an embedding as an input. Moreover, we use five-fold cross validation for direct learners to have the same number of data points in the second phase and to ensure models are evaluated on a held-out dataset.
All the experiments are implemented in Python using PyTorch as a deep learning framework. The experiments are executed on an Ubuntu machine (64GB RAM, one NVIDIA GeForce GPU 8GB) and each experiment is averaged using ten seeds from one to ten and reports one standard error, except for IHDP dataset where results are averaged over 100 runs due to smaller size of the dataset. Final code is available at \textit{\url{https://github.com/jmdvinodjmd/HyperITE}}.

\textbf{HyperITE versions of ITE Learners:} All HyperITE learners have exact same architecture as that of the corresponding baselines, however they are trained through soft weight sharing of hypernet. For HyperTARNet and HyperMitNet, only weights of PO functions are generated and common representation is learned directly from the data, as shows in the figure in main part of the paper. We can generate the entire learner also but that will increase complexity of hypernet and common layers already share information. Similarly, HyperSNet+ generates weights for PO functions and propensity score estimator and latent factors are directly learned. For indirect learners, there is only one step so PO functions are directly generated by hypernet. HyperSLearner and HyperTLearner need to generate one and two networks, respectively. However, direct learners need multiple stages and have multiple models. For HyperDRLearner, in step one, one hypernet is used to generate weights for three networks corresponding to two PO functions and one propensity score estimator, however, second step is similar to DRLearner as there is direct learning of treatment effect from pseudo-outcomes. HyperRALearner needs to generate weights only for PO functions in step one, while step two is similar to RA-Learner as treatment effect is directly regressed on pseudo-outcomes. Number of embedding vectors needed in an HyperITE depends on number of networks whose weights have to be generated, as hypernet need to identify the network, e.g., HyperTLearner needs two embedding vectors corresponding to two PO functions.

\section{ADDITIONAL EXPERIMENTS}
\label{app_experiments}
Here, we provide some additional results related to effect of embedding size, type of hypernet, dataset size on the performance of HyperITE, and convergence of HyperITE against ITE learners, as discussed below.

\subsection{Effect of Hypernets Type / Weight Generation Strategy}
\label{subapp_hypernet}
\begin{figure}[htb!]
	\centering
	\begin{subfigure}[b]{0.49\textwidth}
		\centering
		\includegraphics[width=\textwidth]{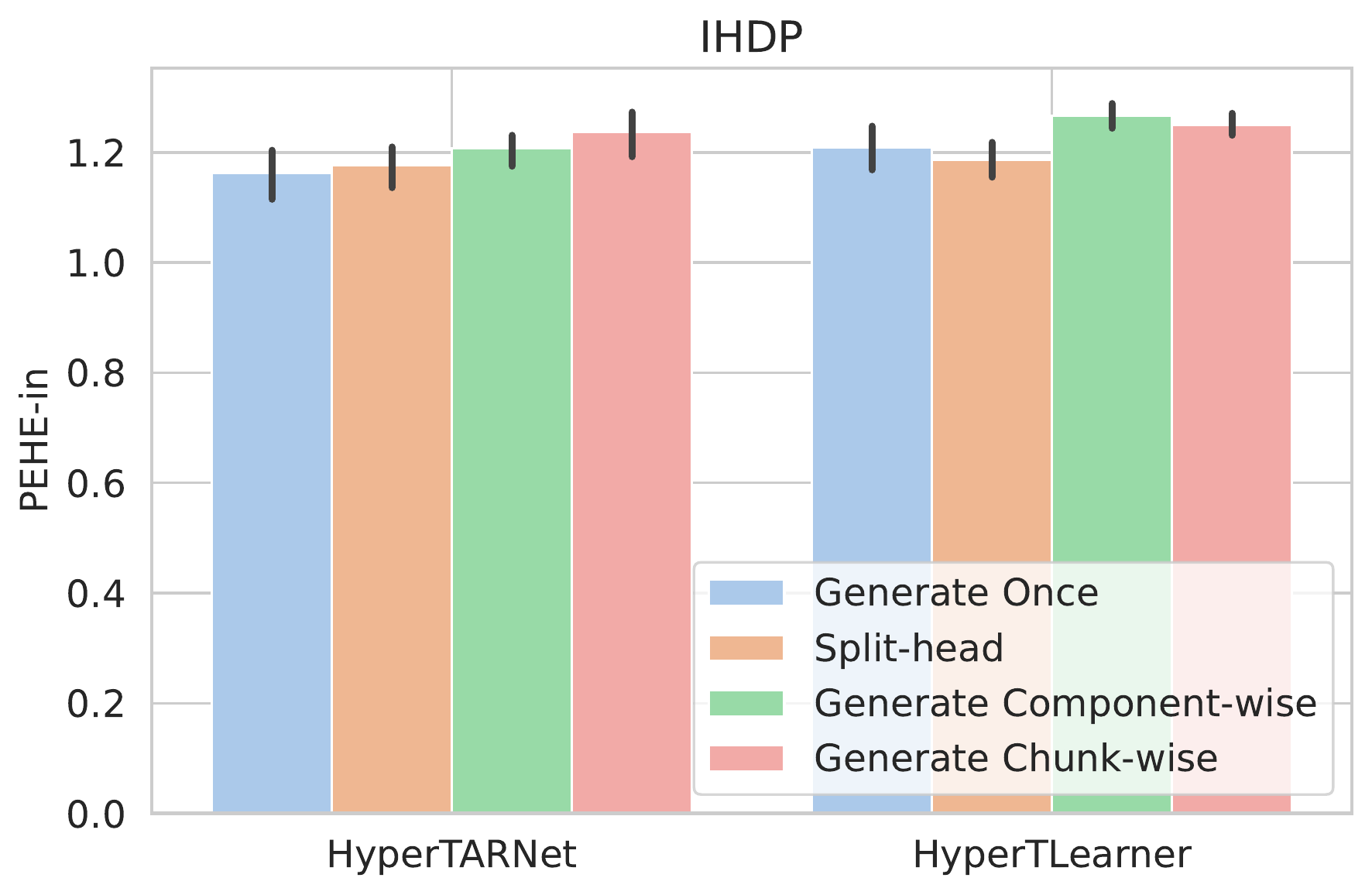}
	\end{subfigure}
	\begin{subfigure}[b]{0.49\textwidth}
		\centering
		\includegraphics[width=\textwidth]{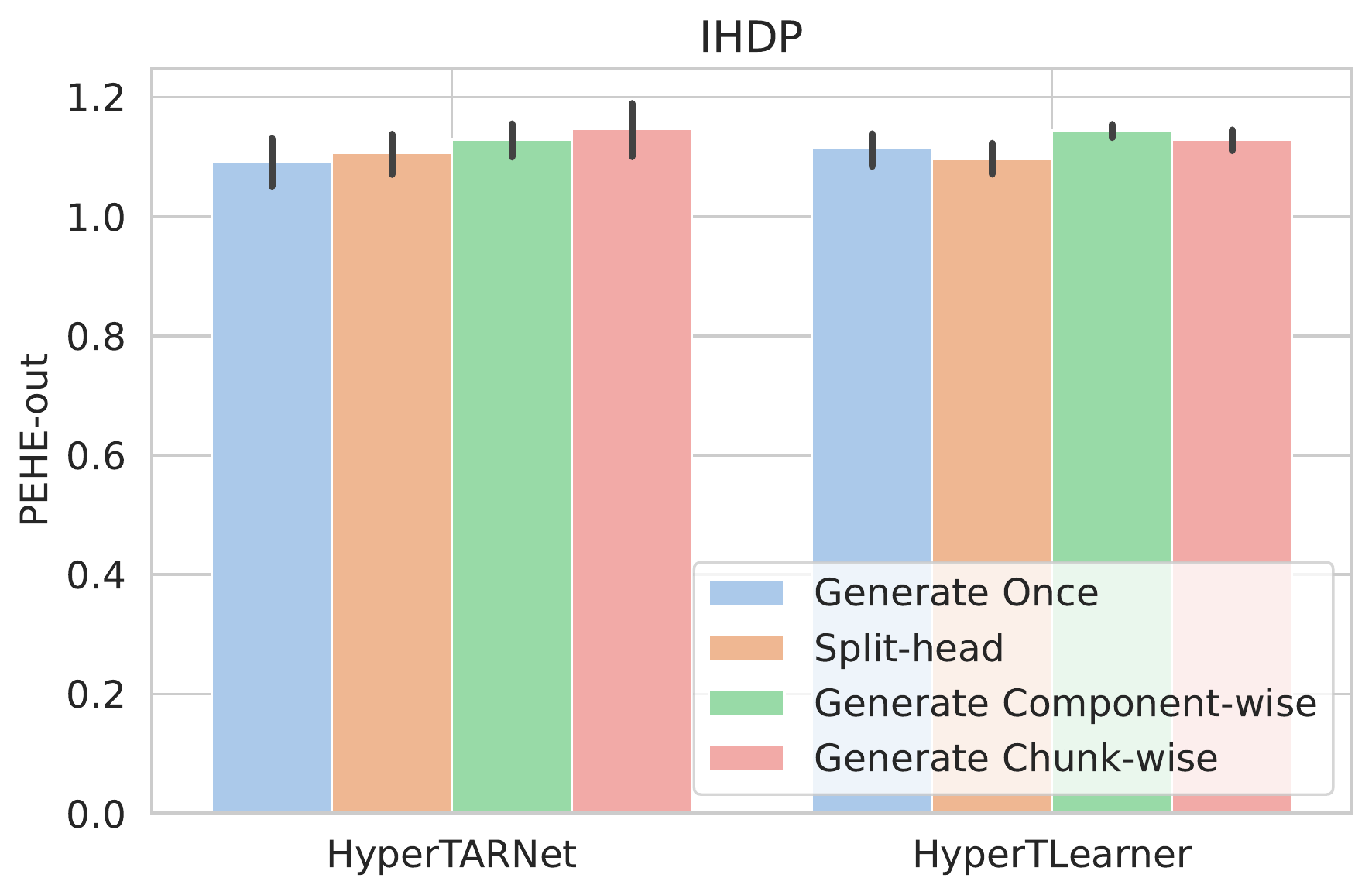}
	\end{subfigure}
	
	\begin{subfigure}[b]{0.49\textwidth}
		\centering
		\includegraphics[width=\textwidth]{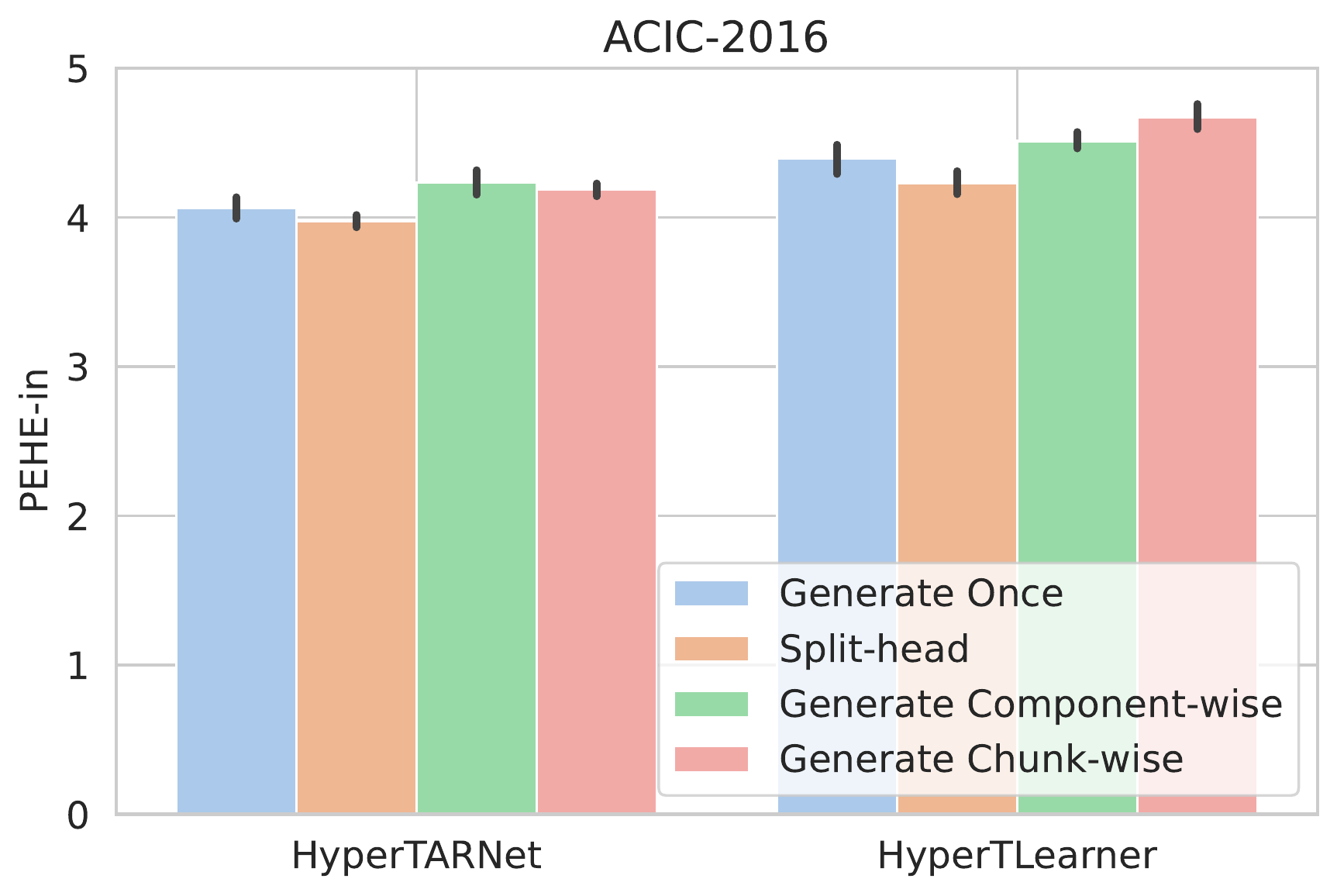}
	\end{subfigure}
	\begin{subfigure}[b]{0.49\textwidth}
		\centering
		\includegraphics[width=\textwidth]{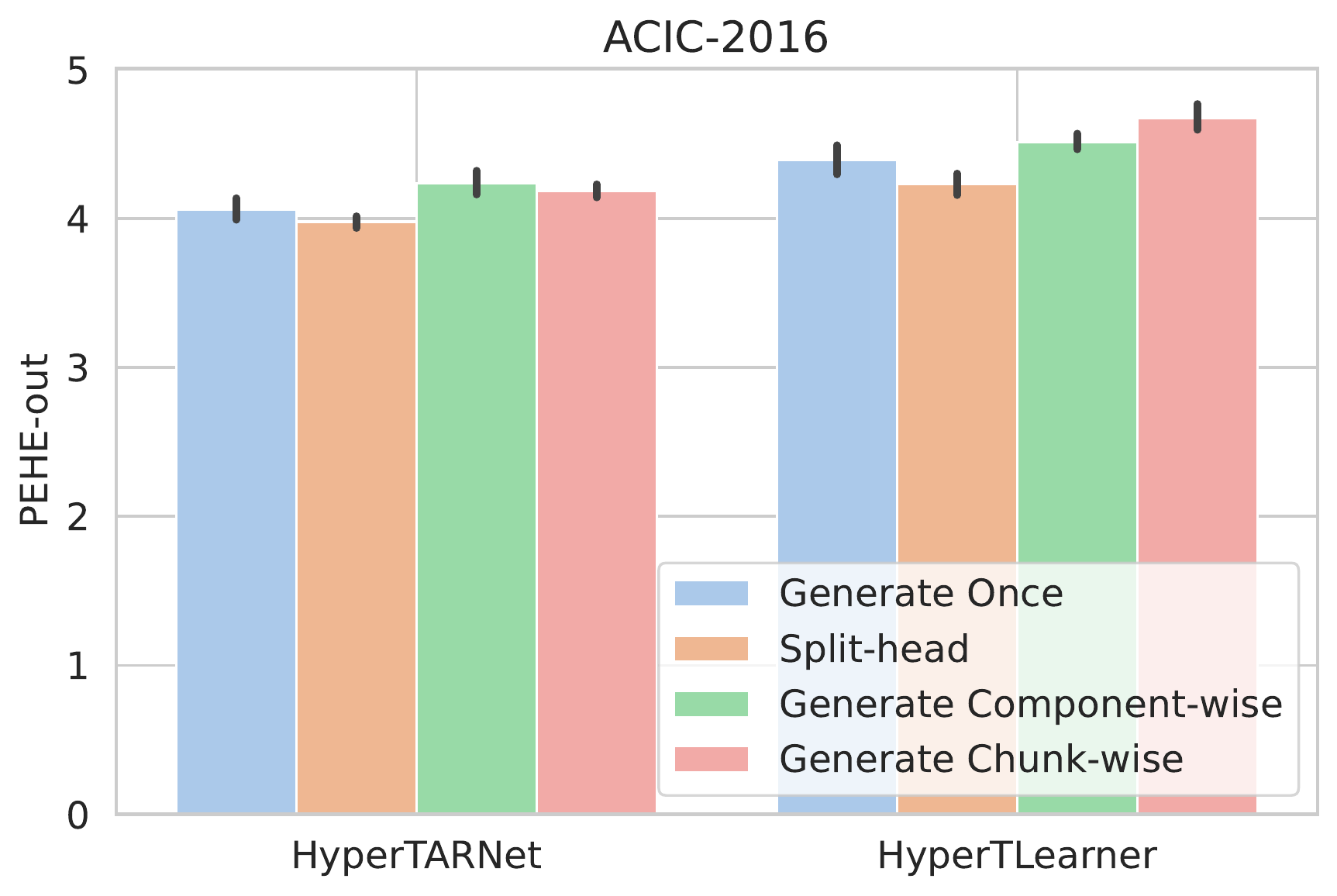}
	\end{subfigure}
	
	\begin{subfigure}[b]{0.49\textwidth}
		\centering
		\includegraphics[width=\textwidth]{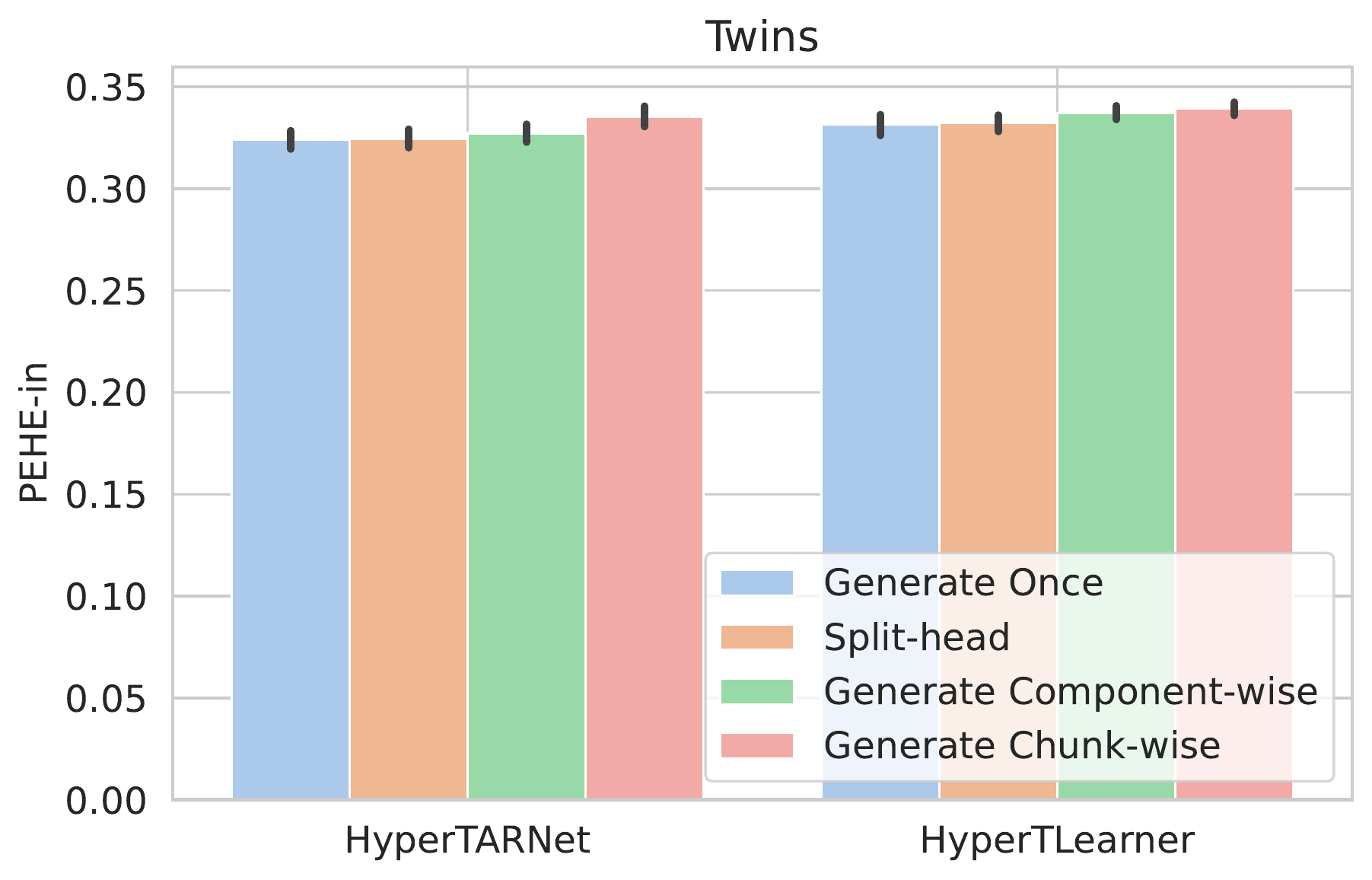}
	\end{subfigure}
	\begin{subfigure}[b]{0.49\textwidth}
		\centering
		\includegraphics[width=\textwidth]{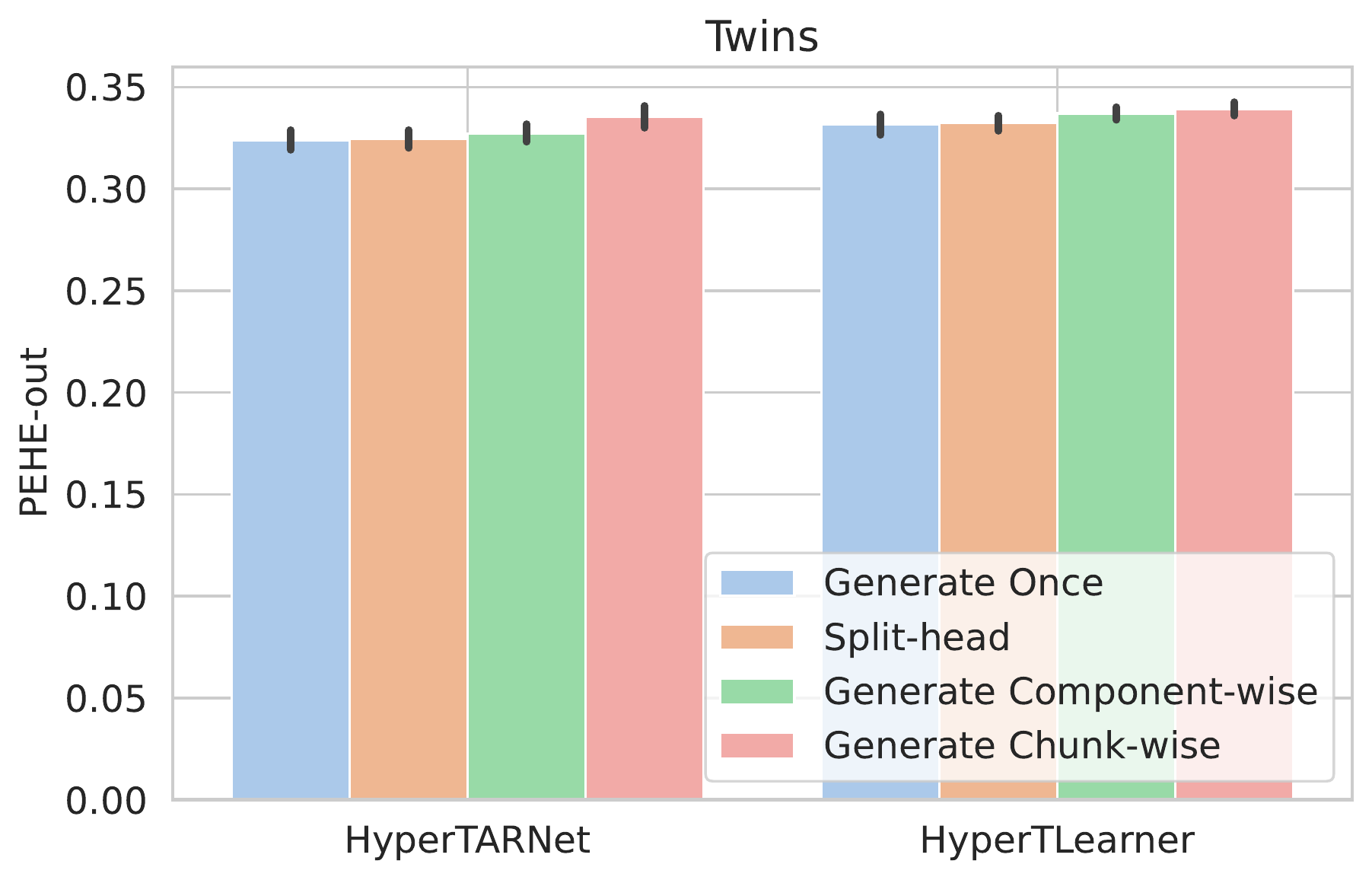}
	\end{subfigure}
	
	\caption{Effect of hypernet's type on performance of HyperTARNet and HyperTLearner using PEHE-in (left) and PEHE-out (right) performance metrics on IHDP, ACIC-2016 and Twins datasets.}
	\label{fig_hypernet_type}
\end{figure}
Hypernets are neural networks that generate weights for ITE learners. Depending on architecture of ITE learner, size of output layer of hypernet can become large and can affect convergence of learning process. To manage complexity of hypernets, which is one of the challenges in hypernets, instead of generating weights once/together, weights can be generated in chunk-wise or component-wise (component refers to a layer or a channel) \cite{Oswald2020Continual}. However, each chunk or component will need an additional embedding so that hypernet could distinguish among them for weight generation. Additionally, split-head approach uses multiple heads for weights generation and can work in combination with other approaches. These target weight generation strategies have their own advantages and disadvantages, e.g., for generate once strategy, size of output layer is highest but its input embedding space is simple and all weights are used and each layer's weight is generated together. On the other hand, chunk-wise generation simplifies output space but makes the input space bit complex, and not all weights are used as weights are generated in fixed chunk sizes. Moreover, if the chunk size is smaller than layer size, then the weights of a layer will not be generated together. Similarly, component-wise weight generation simplifies output space at cost of complicating input space of the hypernets. This approach also ends up not using all the generated weights because weights are generated according to the size of the largest layer of a ITE learner, however, weights of each layer are generated altogether. The split-head weight generation has the best of all approaches as it simplifies the output space without affecting the input space. However, one limitation of the split-head approach is that we can't have a large number of heads for weight generation, otherwise, that will make each head very small.

Fig.~\ref{fig_hypernet_type} studies the effect of type of hypernet on performance of HyperITE on IHDP, ACIC-2016 and Twins datasets with HyperTARNet and HyperTLearner. For chunk-wise generation, we have used ten chunks that reduces output layer size by ten times while increasing the number of embeddings by the same number. However, the effect on output space is huge as compared with input space so we now focus only on output space. Component-wise generation (here referring to layer-wise generation) reduces output space only by 80\% and does not have a large drop as in chunk-wise generation. This is because of having only two hidden layers in our ITE learners. Split-head approach is applied to generate once weight strategy with two splits so it helps to reduce number of parameters in output layer of the hypernet by two times. In terms of performance, from the figure, we observe that generate once and split-head weight generation strategies outperform the rest across all the datasets as well for both HyperITEs. However, there is no clear pattern between chunk-wise and layer-wise approach of weight generation in hypernets for the problem under study. Moreover, there is also not a clear winner between split-head and generate once strategy as split-head performs best on ACIC-2016 dataset while generate once performs best, with a tiny margin against split-head, on Twins dataset. While on IHDP dataset, split-head performs best for HyperTLearner and generate once performs best for HyperTARNet. It is difficult to generalize the choice of hypernet type as it depends on dataset as well as target network.

\subsection{Effect of Embedding Size}
\label{subapp_embedding}
\begin{figure}[htb!]
	\centering
	\begin{subfigure}[b]{0.49\textwidth}
		\centering
		\includegraphics[width=\textwidth]{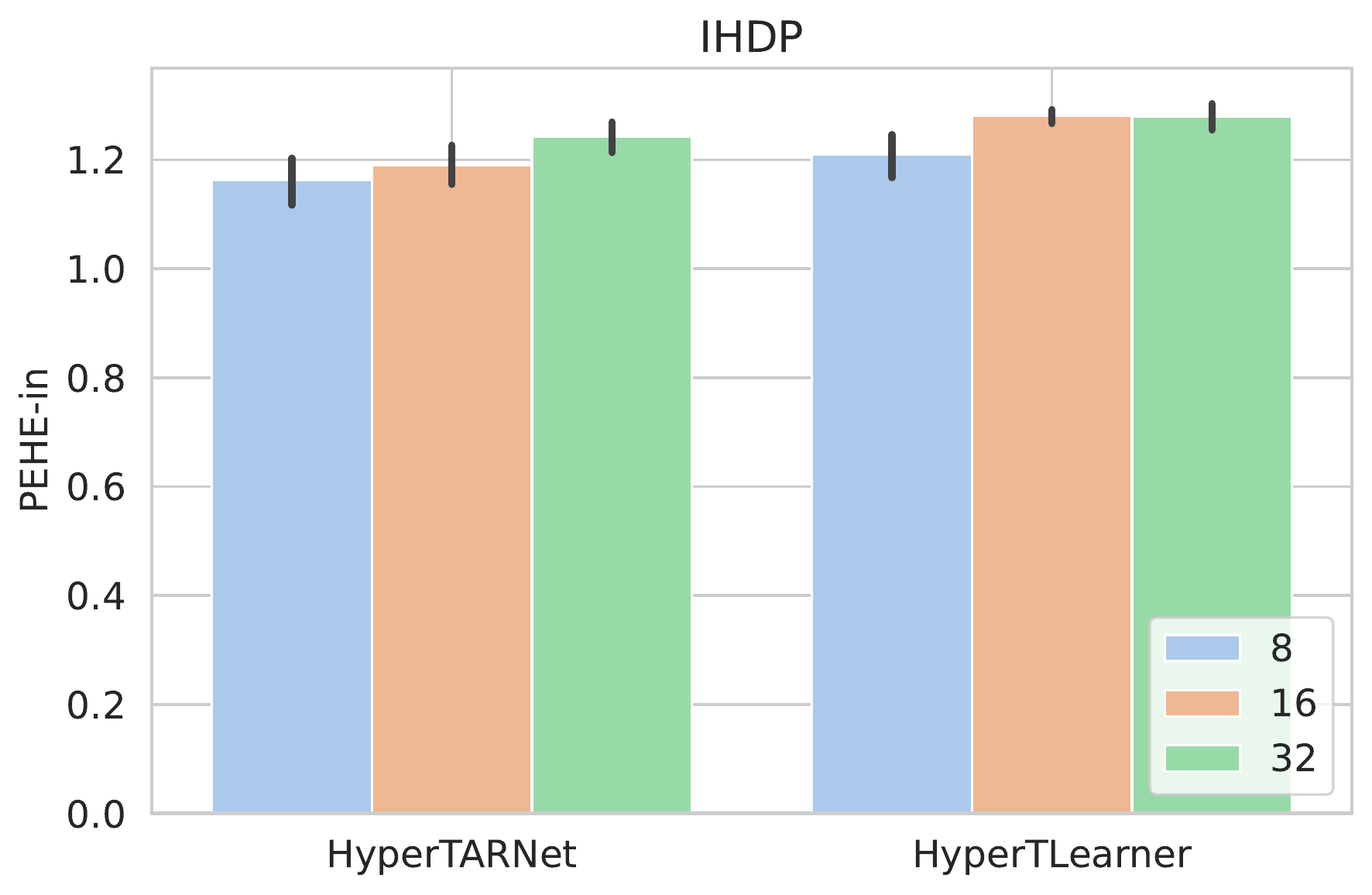}
	\end{subfigure}
	\begin{subfigure}[b]{0.49\textwidth}
		\centering
		\includegraphics[width=\textwidth]{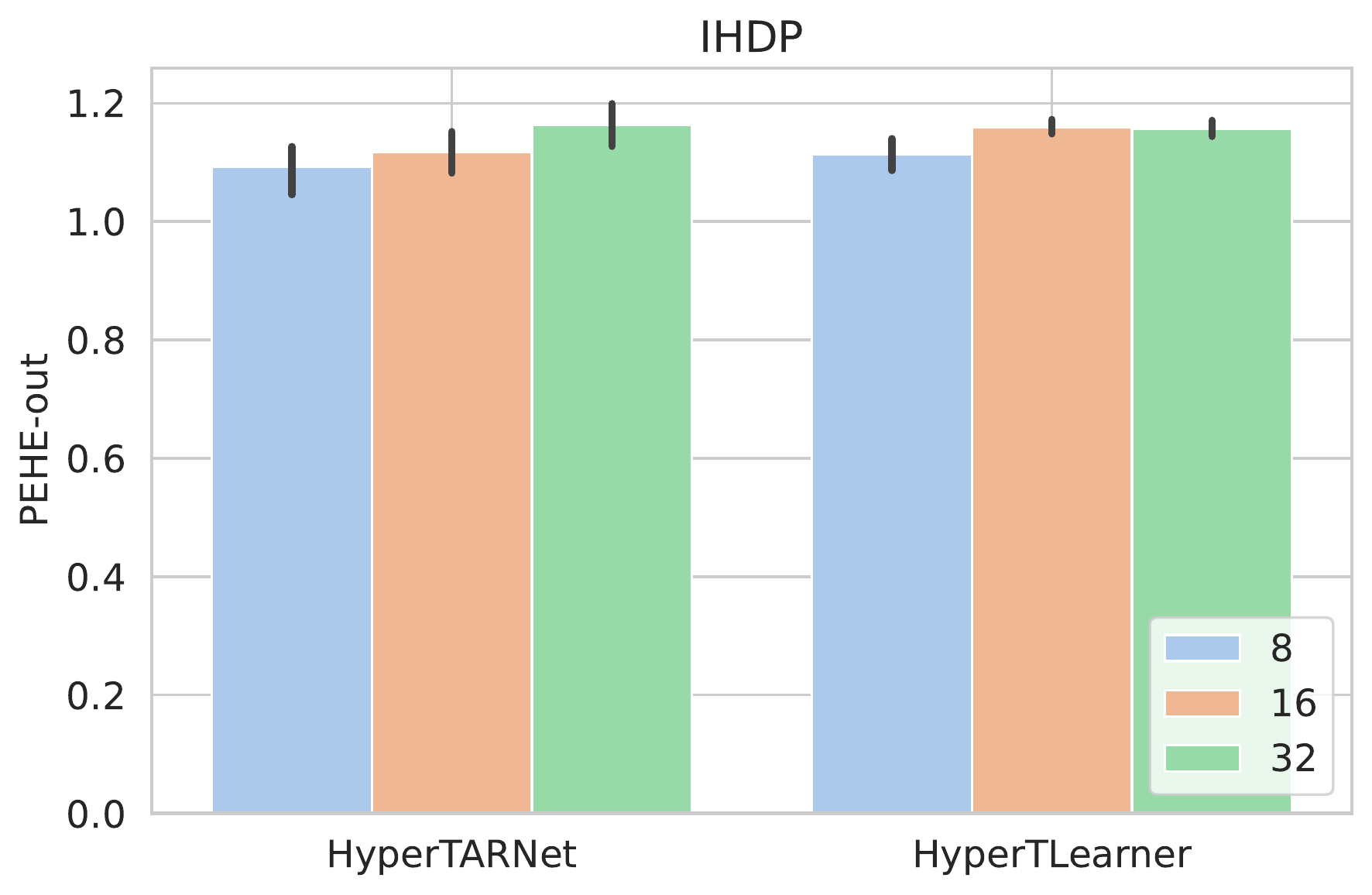}
	\end{subfigure}
	
	\begin{subfigure}[b]{0.49\textwidth}
		\centering
		\includegraphics[width=\textwidth]{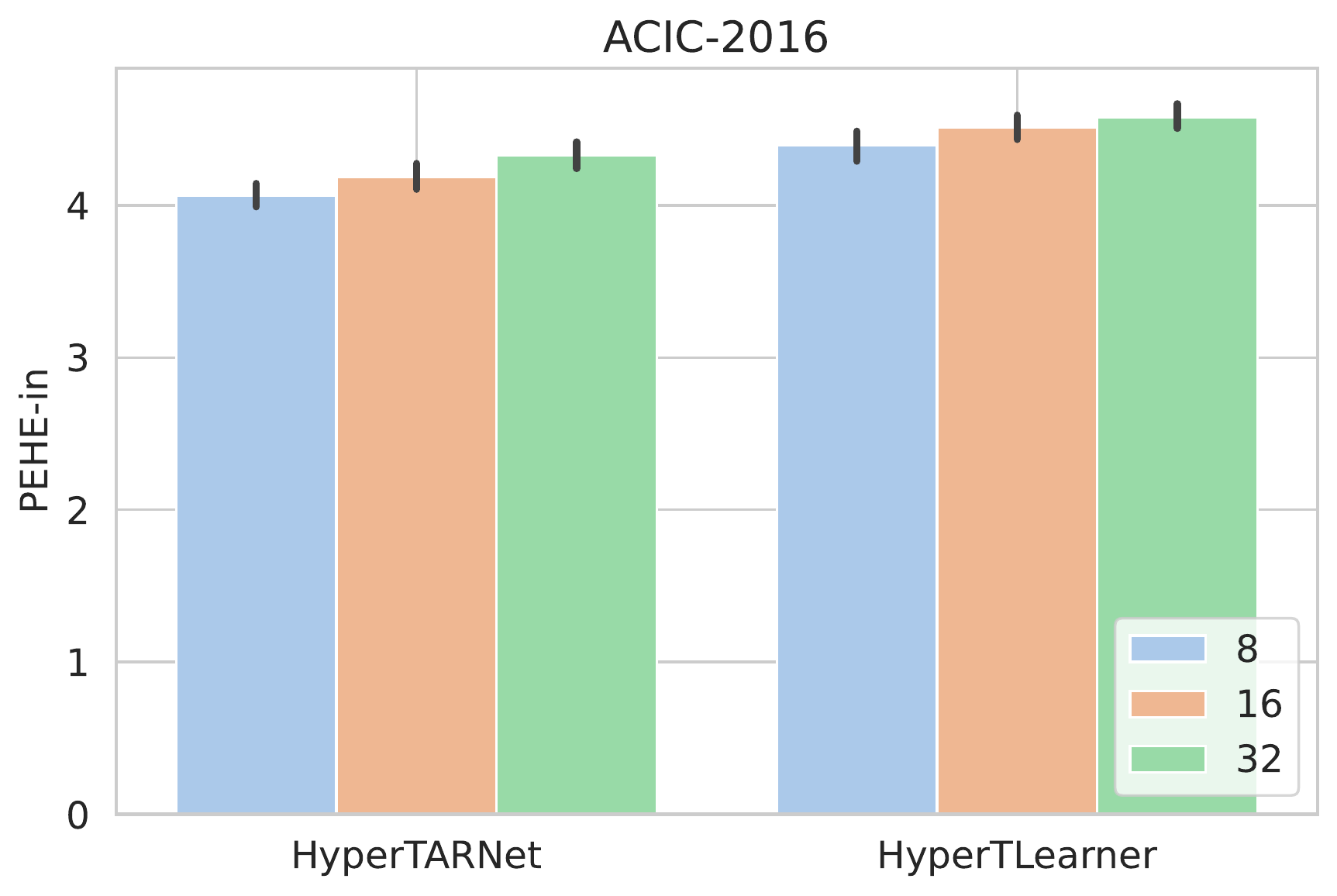}
	\end{subfigure}
	\begin{subfigure}[b]{0.49\textwidth}
		\centering
		\includegraphics[width=\textwidth]{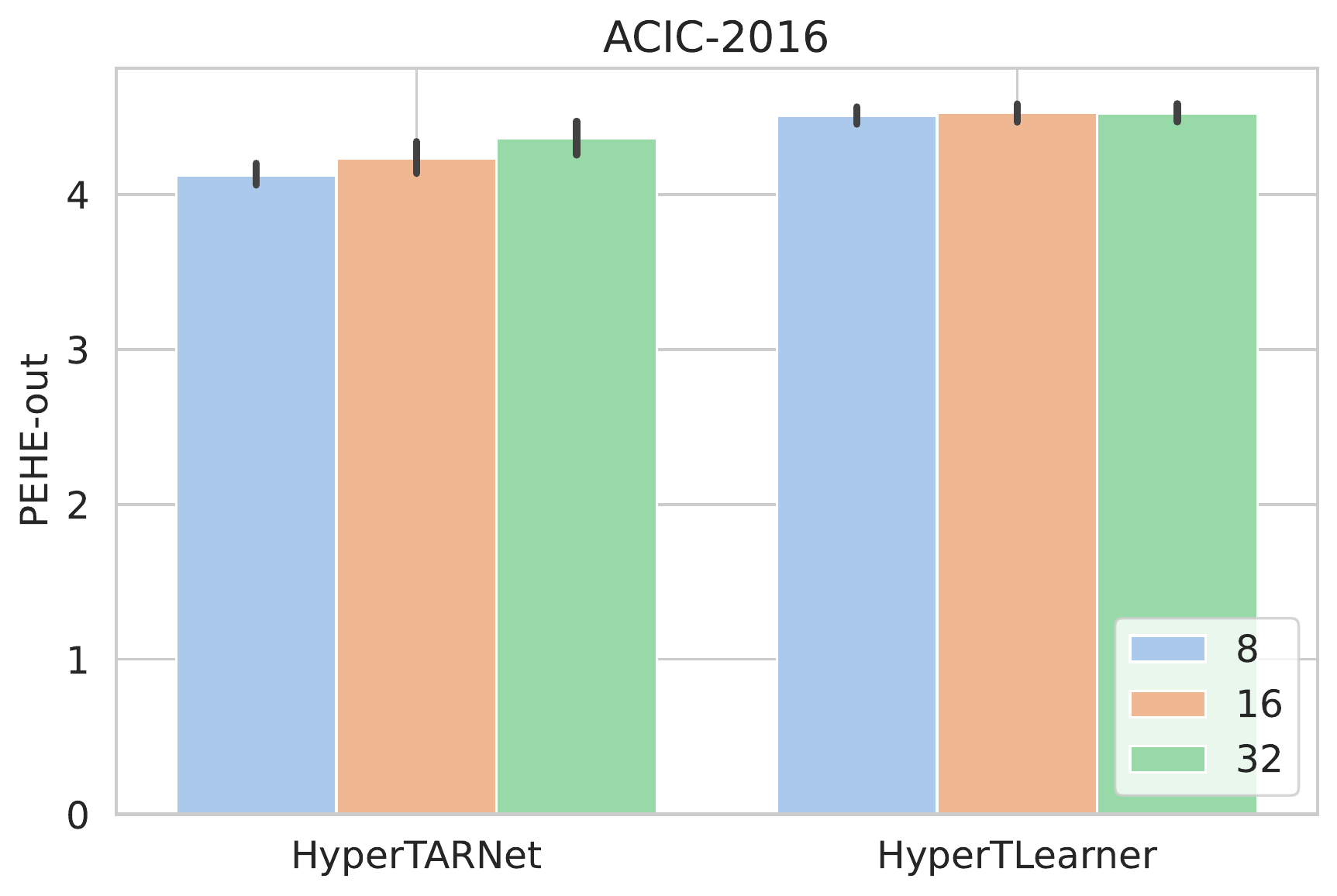}
	\end{subfigure}
	
	\begin{subfigure}[b]{0.49\textwidth}
		\centering
		\includegraphics[width=\textwidth]{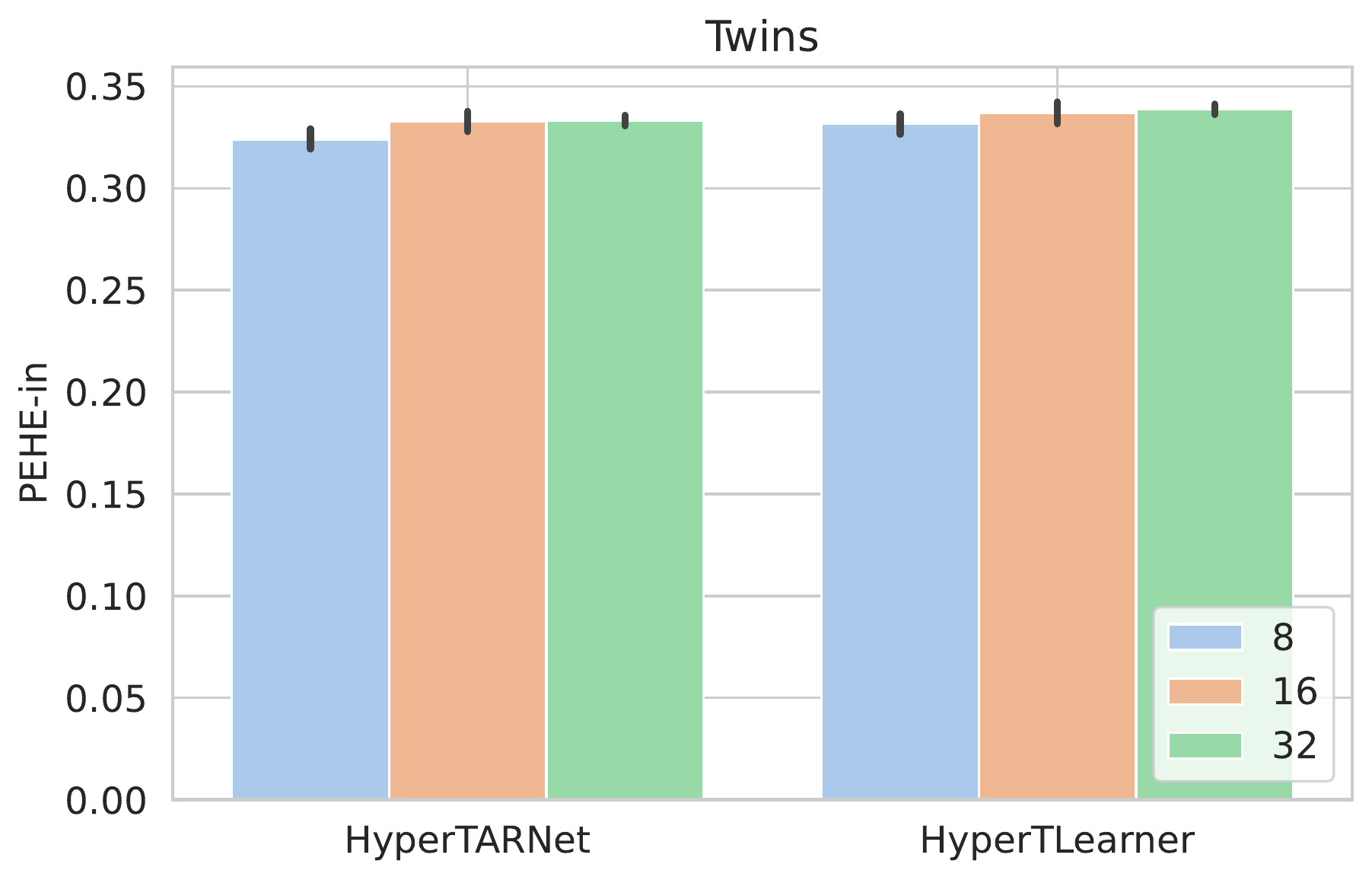}
	\end{subfigure}
	\begin{subfigure}[b]{0.49\textwidth}
		\centering
		\includegraphics[width=\textwidth]{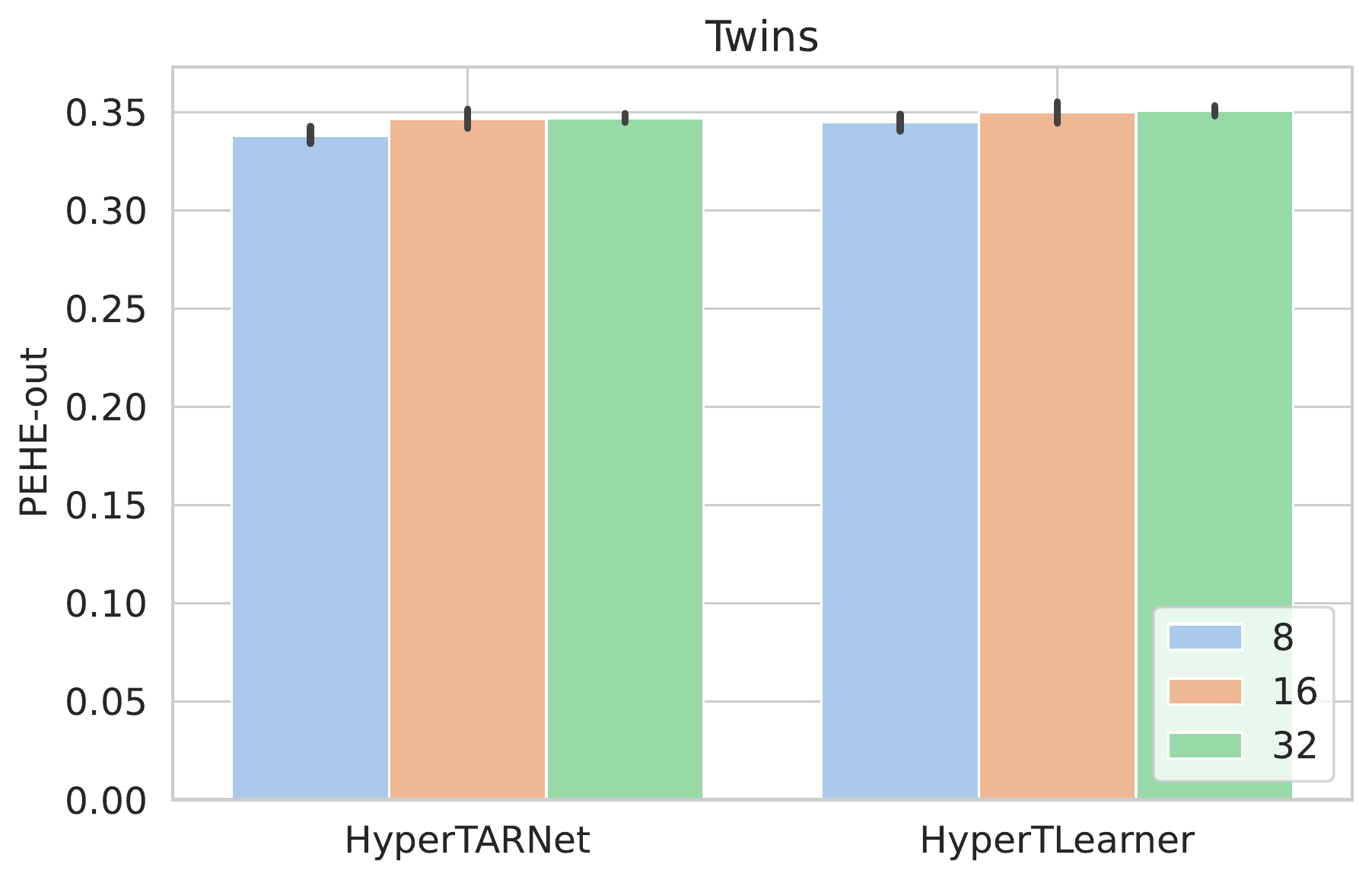}
	\end{subfigure}
	
	\caption{Effect of embedding size on the performance of HyperTARNet and HyperTLearner using PEHE-in and -out performance metrics on IHDP, ACIC-2016 and Twins datasets.}
	\label{fig_embedding}
\end{figure}

Hypernet takes an embedding as an input and maps the embedding to weights for corresponding target network whose weights have to be generated. Fig.~\ref{fig_embedding} presents the effect of embedding size on the performance of the hypernetworks. We have selected one ITE learner from each category, i.e., HyperTARNet from the representation learning and HyperTLearner from meta-learners, and considered embedding sizes of 8, 16 and 32, on IHDP, ACIC-2016 and Twins datasets. From the figure, we observe that HyperTARNet has similar performances on all datasets for both PEHE-in and -out, where embedding of size 8 outperforms the rest. For HyperTLearner also the embedding of size 8 outperforms the rest on both datasets, however, the performance is almost similar for embedding sizes 16 and 32.

\subsection{Comparative Study with Additional ITE Learners}
\label{subapp_additional_ITE}
Here, we discuss tVAE \cite{xue2023assisting} as an additional baseline, which is disentangled representation based variational autoencoder and learns latent representation where first three element corresponds to two PO functions and one to propensity score estimator. So, by architecture, tVAE is a concrete ITE learner that provides end-to-end information sharing. We have used official implementation of tVAE. We present the comparative study here as the results were unstable with ACIC dataset as it leads to NaN values. Results with rest of the datasets are presented in Table~\ref{tab_tvae}, and from the table, we find that it performs worse than FlexTENet \cite{curth2021inductive} which is another concrete ITE learner with flexible architecture for information sharing. Overall, tVAE performs the worst on IHDP, however, it performs better on T-Learner and RA-Learner on Twins dataset.

\begin{table}[htb!]
	\centering
	\caption{Comparative study of ITE learners against HyperITE learners on IHDP and Twins benchmarks using PEHE-out (lower is better) as a performance metric. Each experiment is averaged over multiple runs and values in parenthesis refer to one standard error.}
	\label{tab_tvae}
	\begin{tabular}{lrrrrrr}\\ \hline
		\multicolumn{1}{c}{\multirow{2}{*}{\textbf{Learner}}} &
		\multicolumn{1}{c}{\textbf{IHDP}} &
		\multicolumn{1}{c}{\textbf{Twins}} \\
		\multicolumn{1}{c}{} &
		\multicolumn{1}{c}{\textbf{PEHE-out}} &
		\multicolumn{1}{c}{\textbf{PEHE-out}} \\ \hline
		FlexTENet      & 1.19 (0.011) & 0.328	(0.000)  \\
		tVAE           & 1.38 (0.003) & 0.366	(0.001)  \\ \hline
		SLearner       & 0.97 (0.001) & 0.331	(0.001)\\
		HyperSLearner  & \textbf{0.95 (0.001)} &  \textbf{0.324	(0.001)}\\ \hline
		TLearner       & 1.32 (0.003) &  0.410	(0.001) \\
		HyperTLearner  & \textbf{1.12 (0.001)} &  \textbf{0.345	(0.003)} \\ \hline
		DRLearner      & 1.12 (0.002) & \textbf{0.323 (0.001)}\\
		HyperDRLearner & \textbf{1.09 (0.002)} & \textbf{0.323	(0.001)}\\ \hline
		RALearner      & 1.25 (0.002) & 0.384	(0.001) \\
		HyperRALearner & \textbf{1.06 (0.002)} & \textbf{0.329	(0.001)} \\ \hline
		TARNet         & 1.24 (0.002) &	0.366	(0.001) \\
		HyperTARNet    & \textbf{1.08 (0.002)}  &	\textbf{0.338	(0.002)} \\ \hline
		MitNet         & 1.25 (0.004) &	0.324	(0.012) \\
		HyperMitNet  & \textbf{1.12	(0.005)} & \textbf{0.323	(0.012)} \\ \hline
		SNet+          & 1.22 (0.001) &  0.353	(0.003)\\
		HyperSNet+     & \textbf{1.17 (0.001)} &  \textbf{0.325	(0.001)}\\ \hline
	\end{tabular}
\end{table}

\subsection{Effect of Limited Observational Datasets}
\label{subapp_dataset_size}
\begin{figure}[htb!]
	\centering
	\begin{subfigure}[b]{0.47\textwidth}
		\centering
		\includegraphics[width=\textwidth]{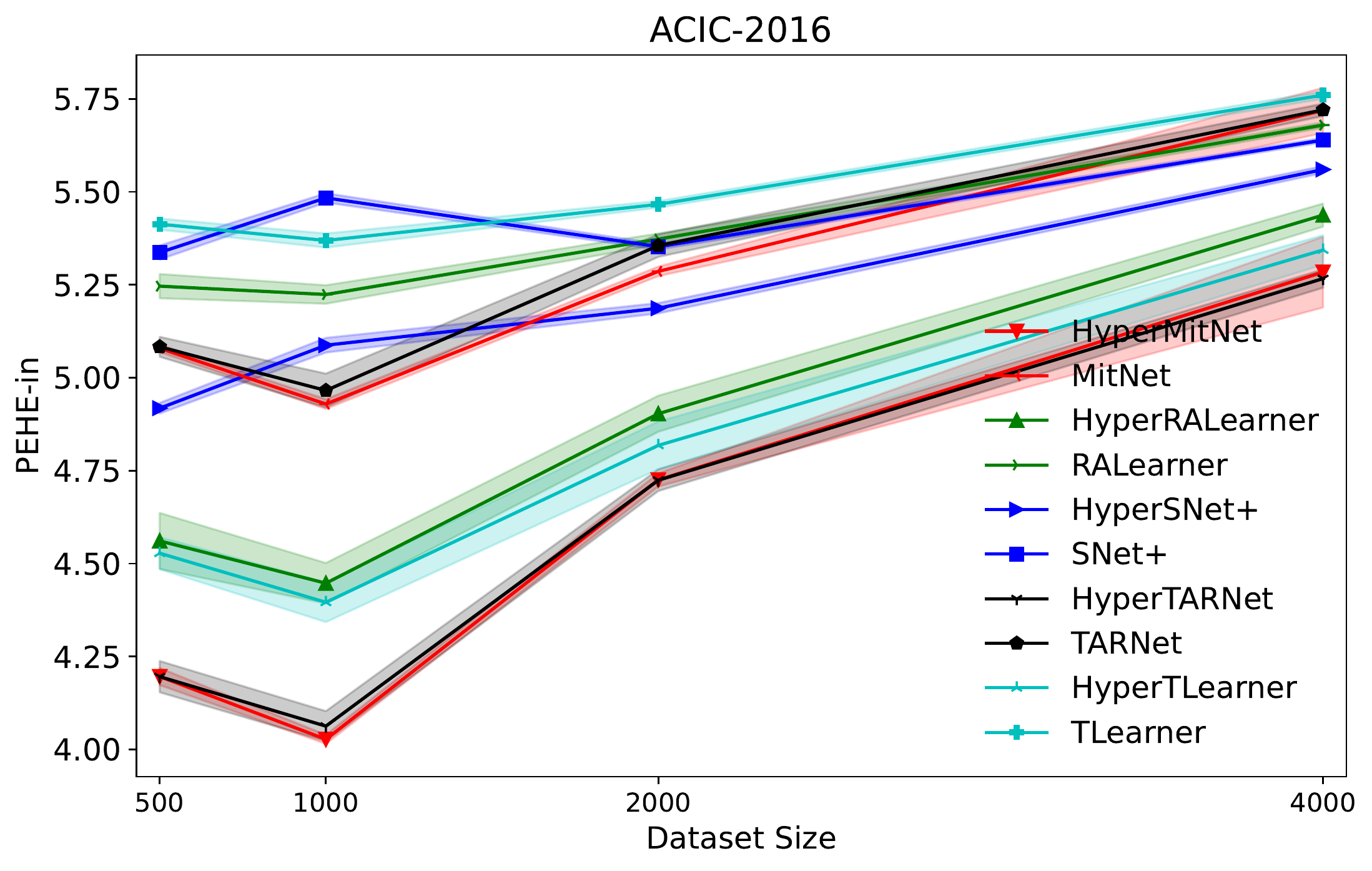}
	\end{subfigure}
	\begin{subfigure}[b]{0.47\textwidth}
		\centering
		\includegraphics[width=\textwidth]{scale-ACIC-2016-2out.pdf}
	\end{subfigure}
	
	\begin{subfigure}[b]{0.47\textwidth}
		\centering
		\includegraphics[width=\textwidth]{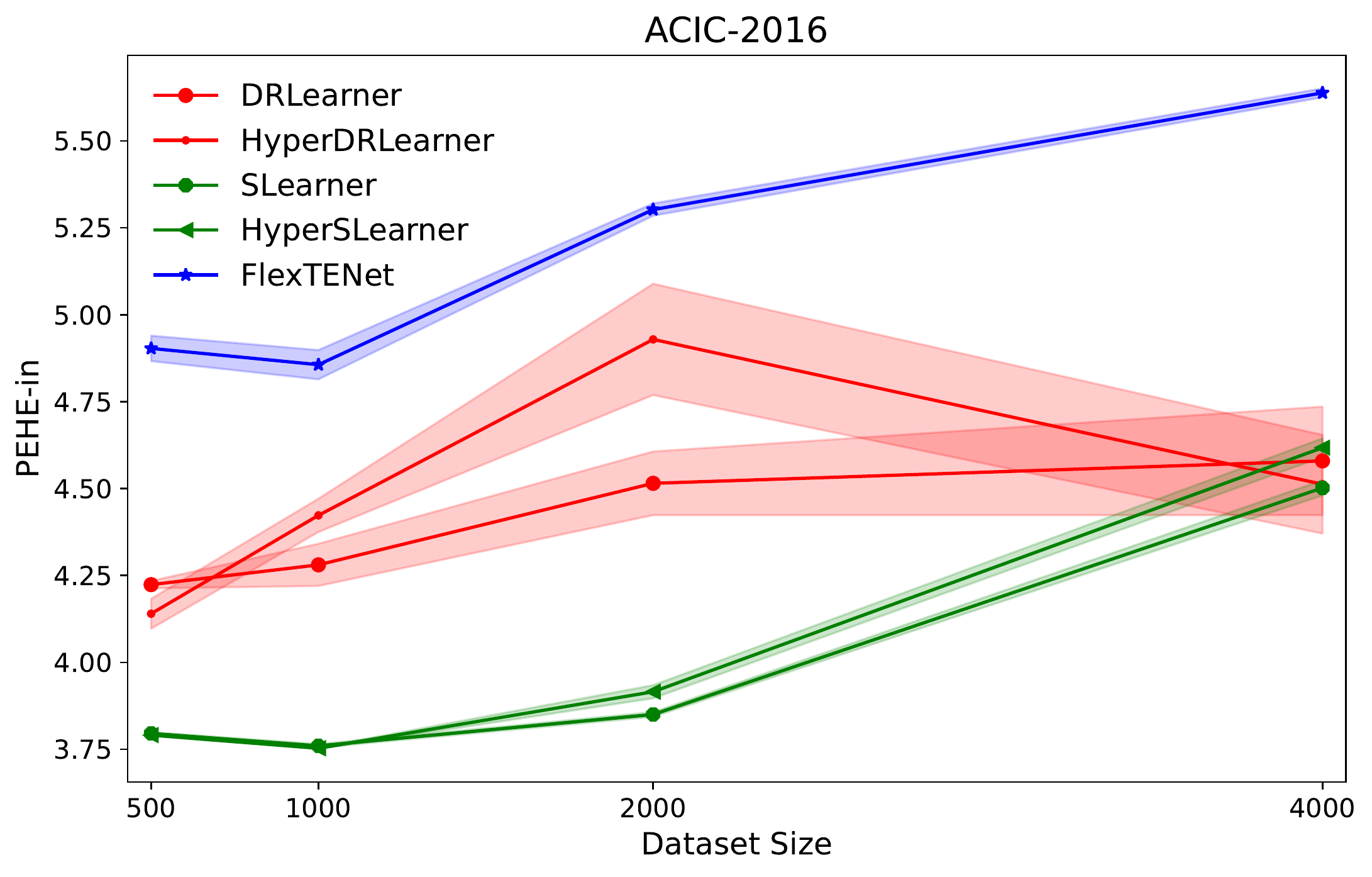}
	\end{subfigure}
	\begin{subfigure}[b]{0.47\textwidth}
		\centering
		\includegraphics[width=\textwidth]{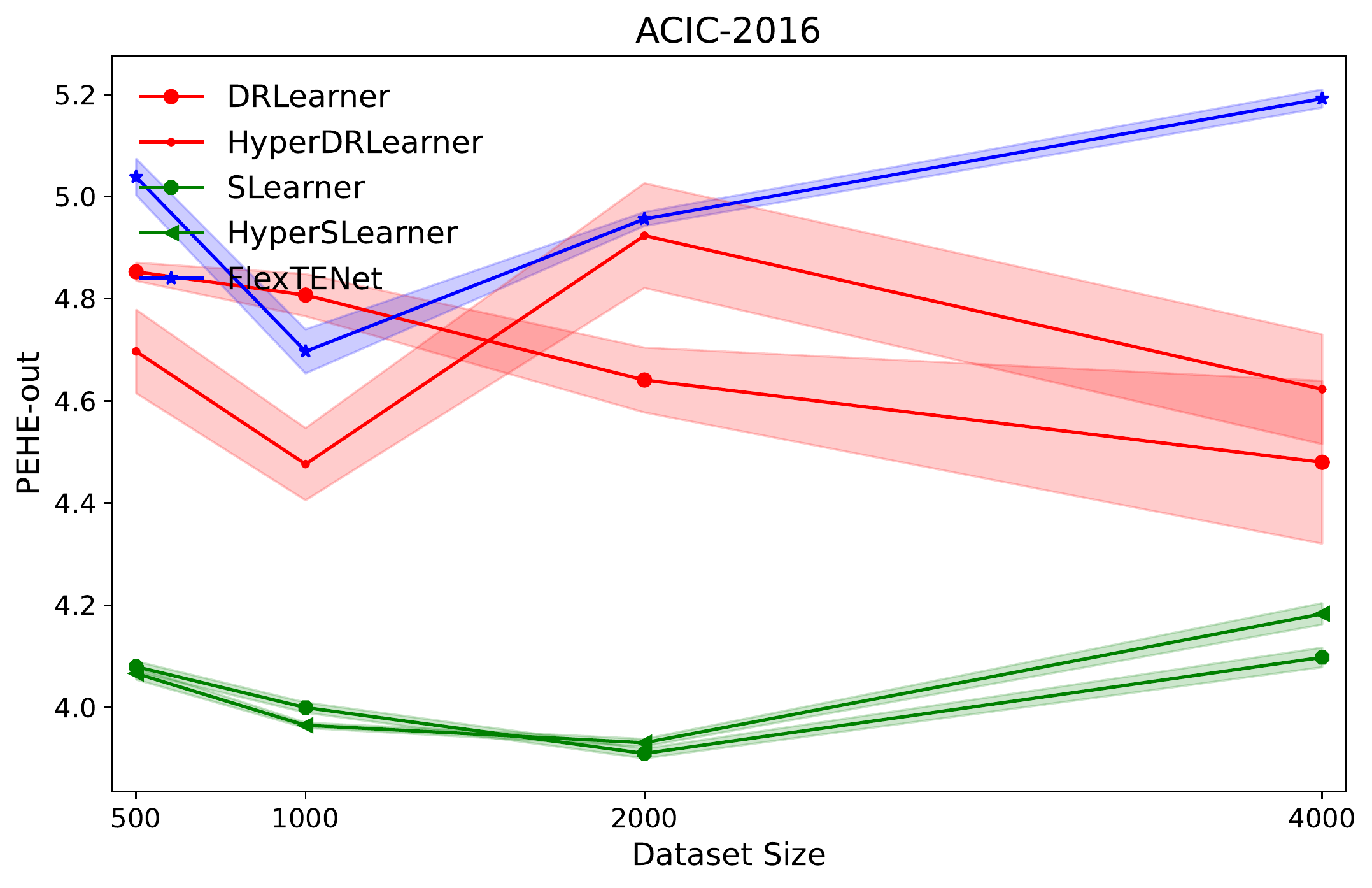}
	\end{subfigure}
	
	\begin{subfigure}[b]{0.47\textwidth}
		\centering
		\includegraphics[width=\textwidth]{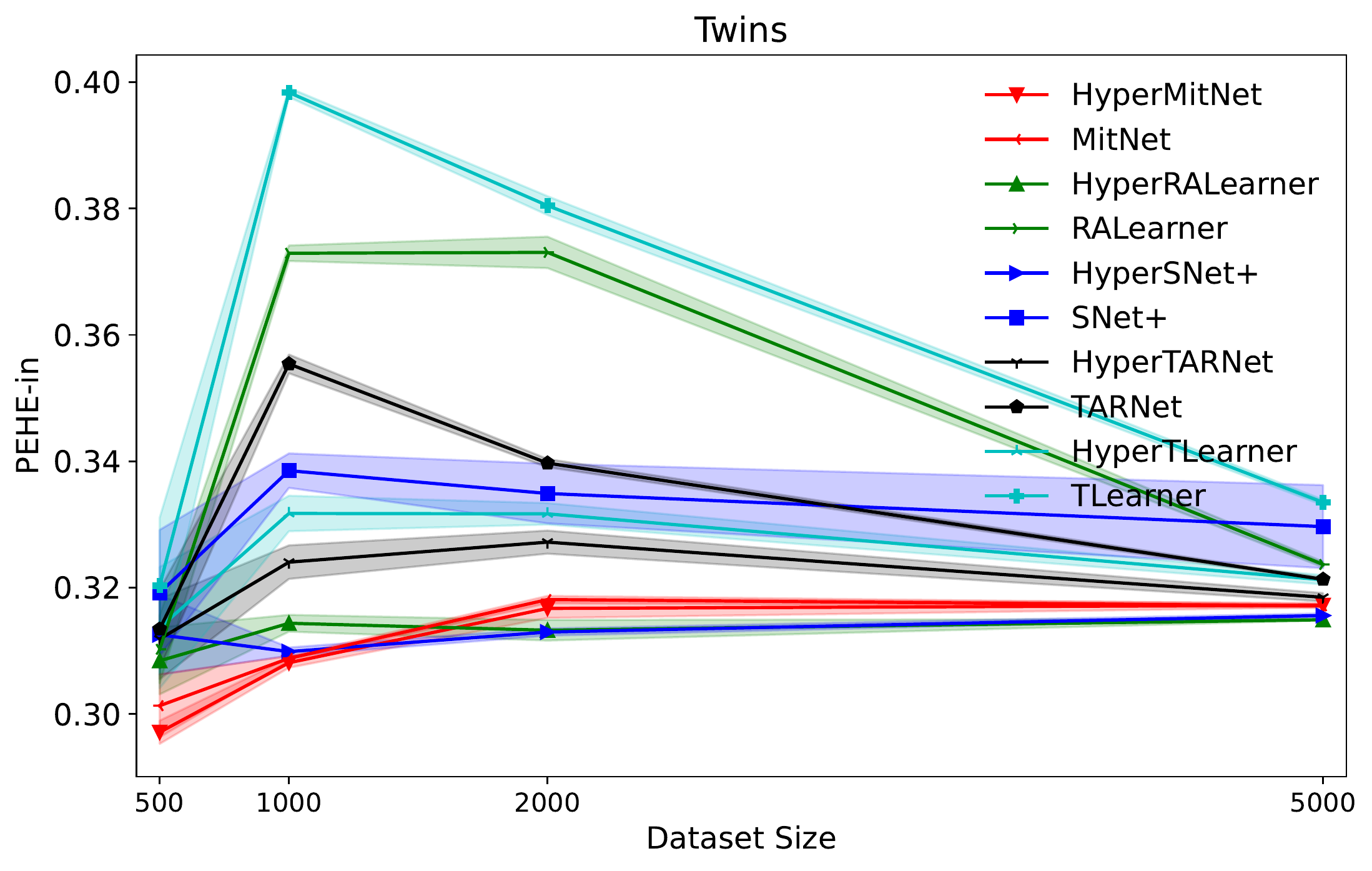}
	\end{subfigure}
	\begin{subfigure}[b]{0.47\textwidth}
		\centering
		\includegraphics[width=\textwidth]{scale-Twins-2out.pdf}
	\end{subfigure}
	
	\begin{subfigure}[b]{0.47\textwidth}
		\centering
		\includegraphics[width=\textwidth]{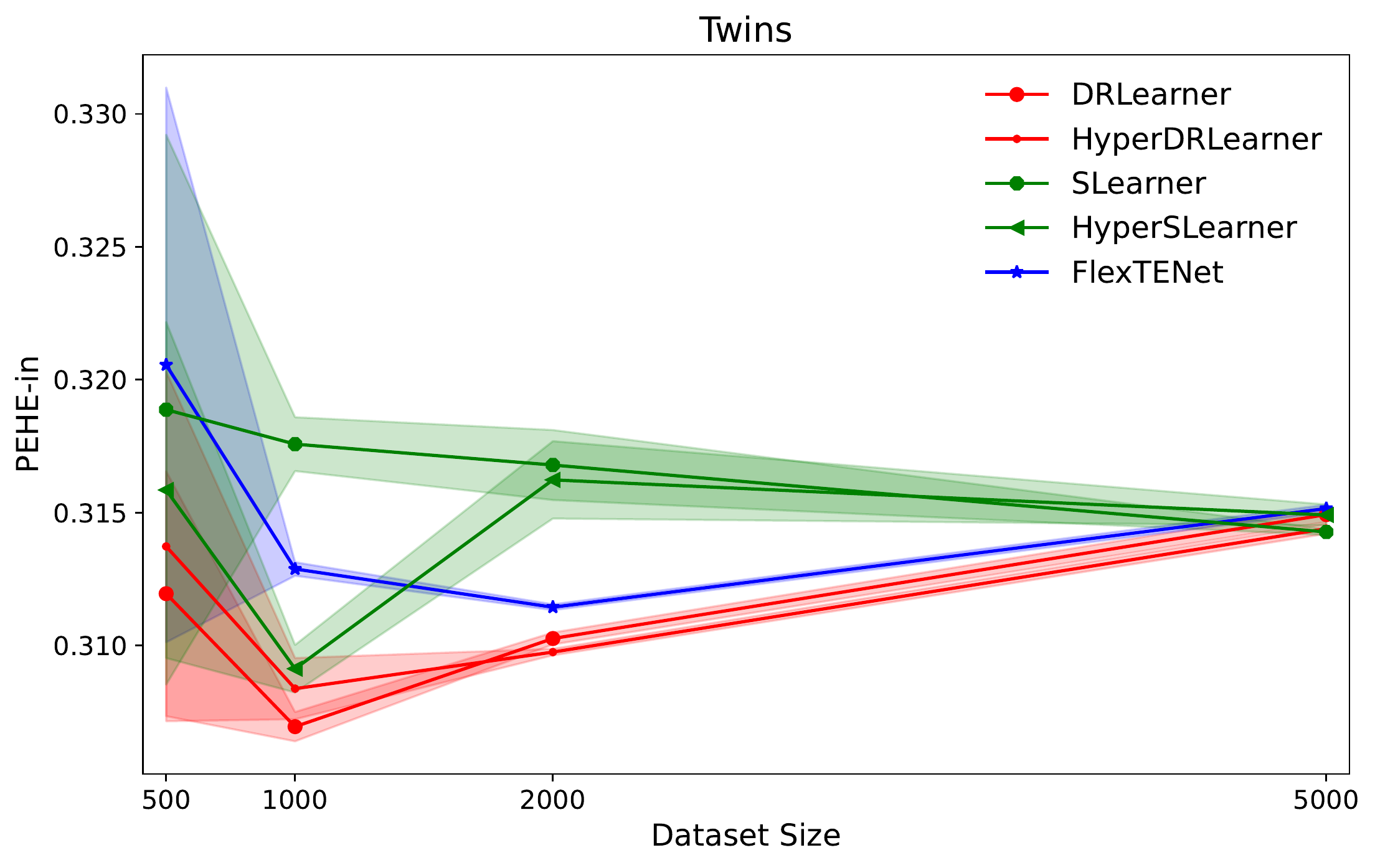}
	\end{subfigure}
	\begin{subfigure}[b]{0.47\textwidth}
		\centering
		\includegraphics[width=\textwidth]{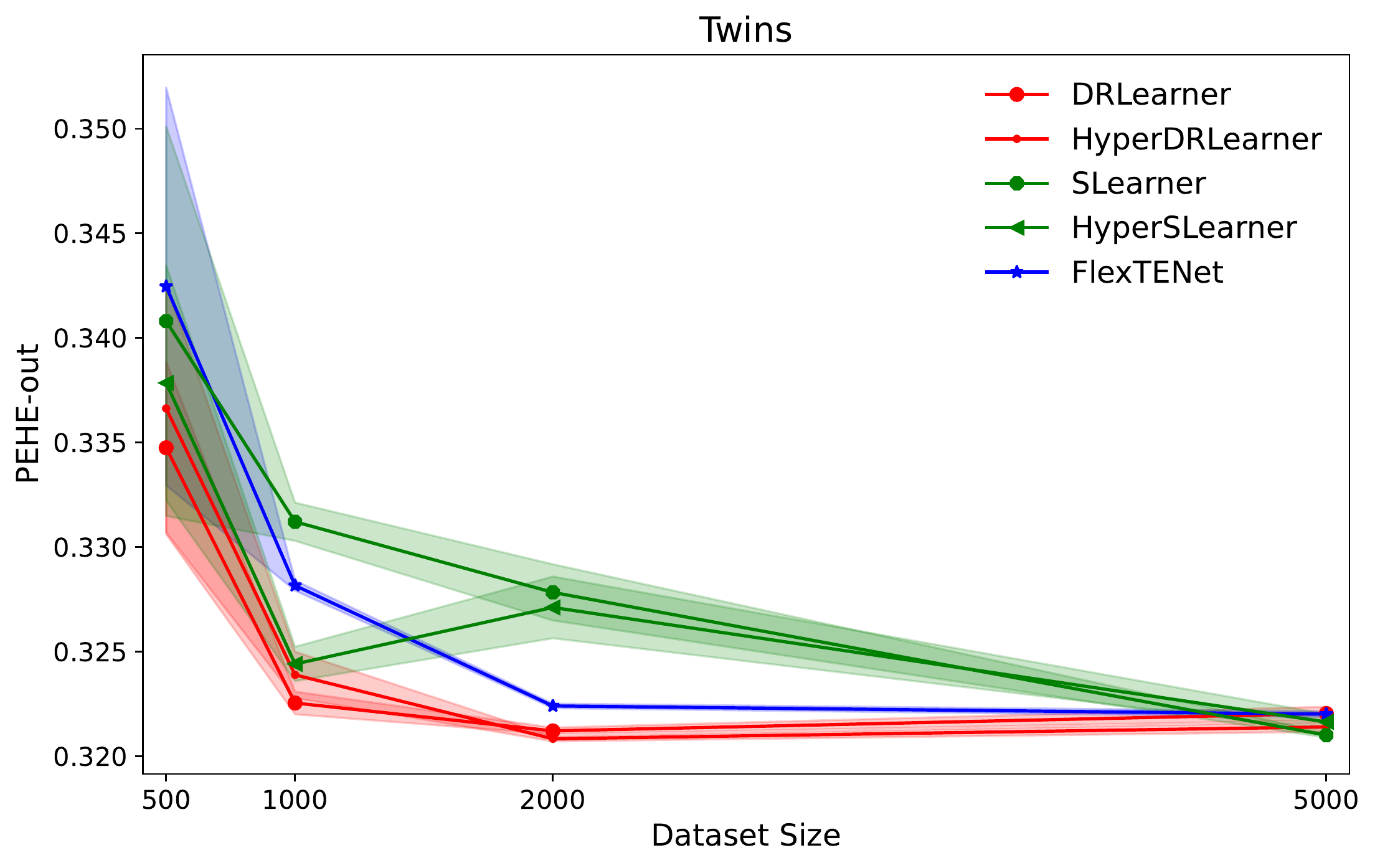}
	\end{subfigure}
	
	\caption{Effect of dataset size on the performance of learners using ACIC-2016 and Twins datasets where left panel shows results for PEHE-in and right panel shows results for PEHE-out (shaded region shows one standard error).}
	\label{fig_dataset_size_all}
\end{figure}

Here, we discuss the effect of dataset size on the performance of the HyperITE against the baselines for PEHE-in as well as for the learners which were not covered in the main part of the paper and were also not showing consistent performance. Moreover, for the sake of completeness, we re-include the results presented earlier so that all the learners can be studied in one place.
We present results in Fig.~\ref{fig_dataset_size_all}. From the figure, we observe similar performance of learners on PEHE-in and -out performance metrics, and as discussed earlier, results show that HyperITEs show improvements over the baselines and results become better with decreasing dataset size.
For HyperSLearner on both the datasets, there is not any clear pattern, however, it shows some improvements with smaller sizes. This is potentially due to having single learner and hence end-to-end information sharing. While, as observed earlier, HyperDRLearner does not show any clear pattern, and this is potentially due to regularization of second step by errors in the first step. Moreover, FlexTENet also does not show consistent results over the two datasets and with a decrease in dataset size the performance drops for Twins and shows unclear performance with ACIC-2016.

\subsection{Convergence of ITE and HyperITE Learners}
\label{subapp_convergence}

\begin{figure}[htb!]
	\centering
	\begin{subfigure}[b]{0.49\textwidth}
		\centering
		\includegraphics[width=\textwidth]{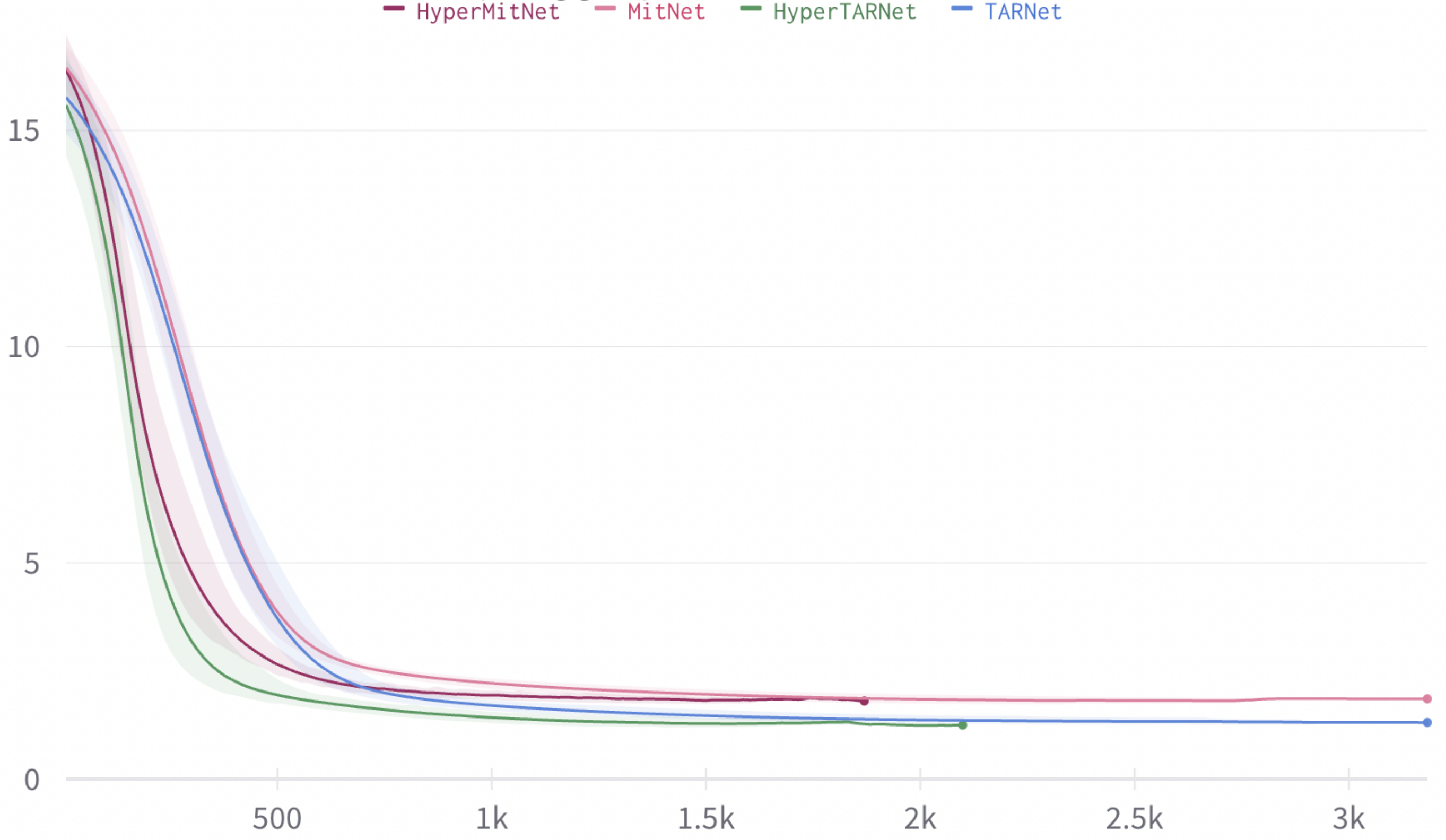}
	\end{subfigure}
	\begin{subfigure}[b]{0.49\textwidth}
		\centering
		\includegraphics[width=\textwidth]{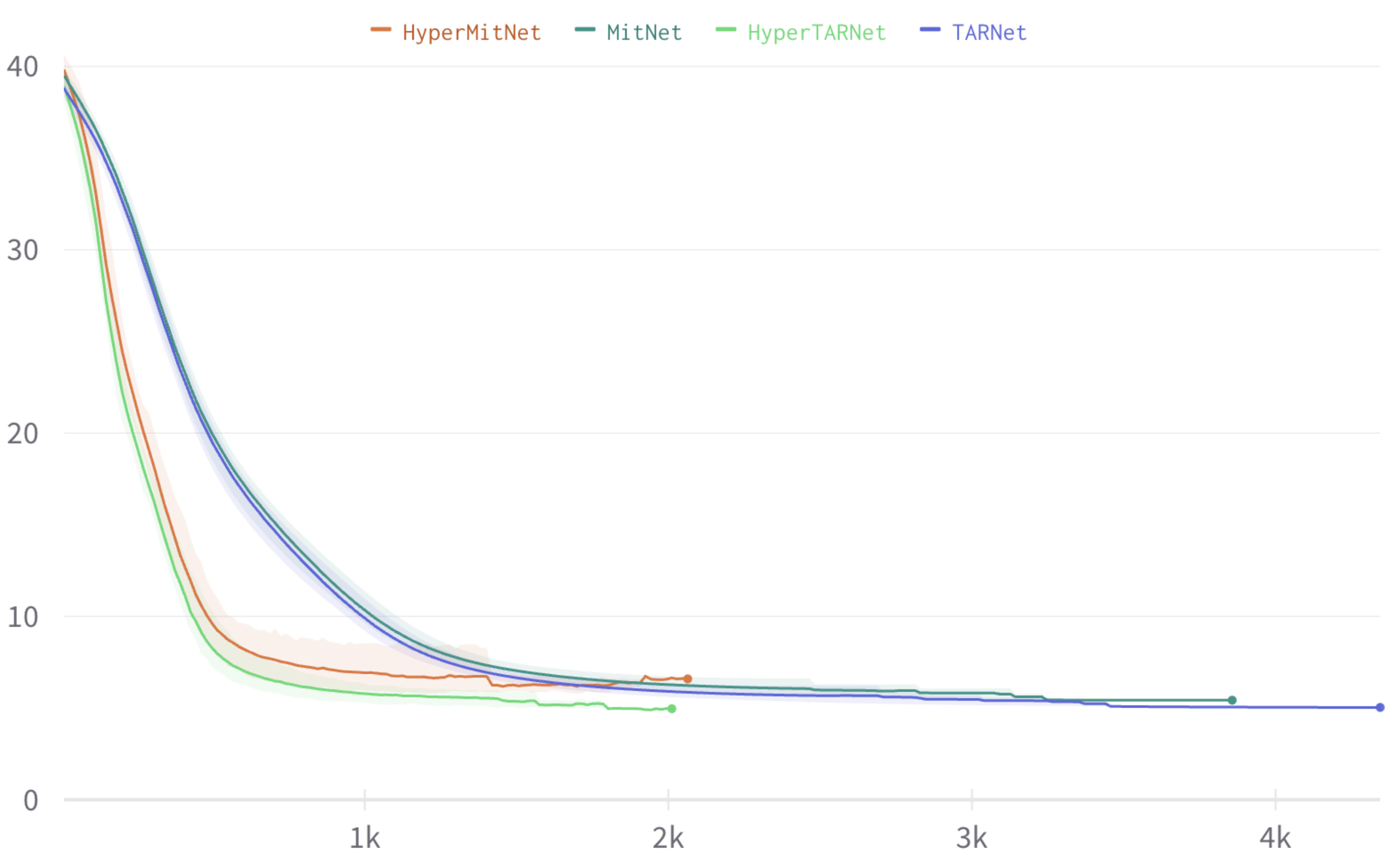}
	\end{subfigure}
	
	\begin{subfigure}[b]{0.49\textwidth}
		\centering
		\includegraphics[width=\textwidth]{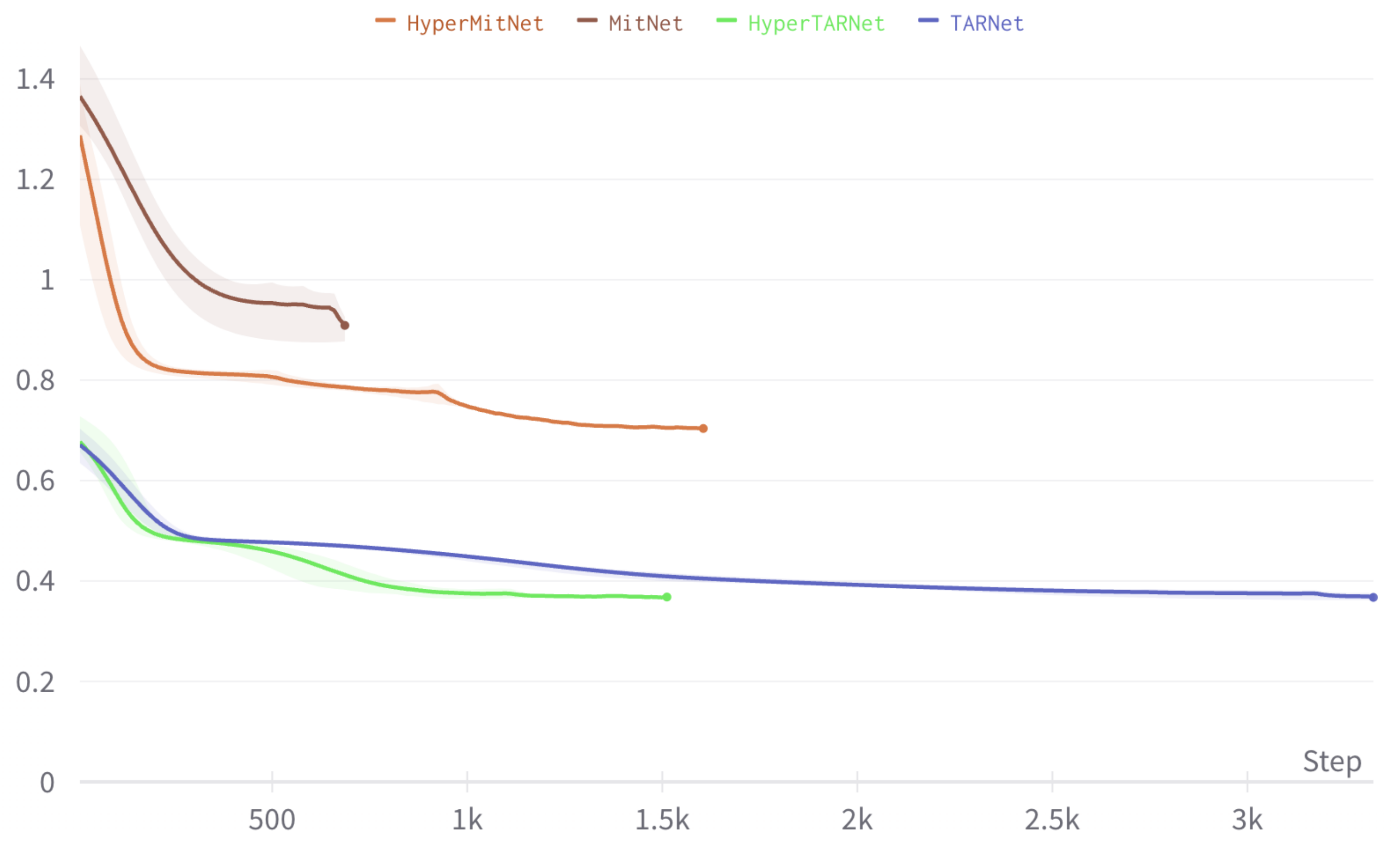}
	\end{subfigure}
	
	\caption{Convergence for validation loss against weight update steps for HyperITEs and corresponding baselines on IHDP (top left), ACIC-2016 (top right) and Twins (bottom) benchmarks (shaded region is difference of max and min values).}
	\label{fig_conv}
\end{figure}

Here, we discuss convergence of HyperITE against ITE learners on three benchmark datasets. However, it is to be noted that we can't compare their convergence for all learners because some learners have multiple training stages where each stage may have multiple models trained independently. But HyperITE can train all models of each stage together through soft weight sharing, i.e., through one model. For example, meta-learners like DR-Learner has two stages where first stage trains three independent models, and T-Learner trains two independent models, whereas HyperDRLearner and HyperTLearner train one model for first stage. 
Fig.~\ref{fig_conv} presents convergence for validation loss for MitNet and TARNet against HyperMitNet and HyperTARNet learners. From the figure, it is clear that HyperITEs converge faster than corresponding baselines across all the benchmarks because of dynamic end-to-end information sharing between potential outcome functions. This helps HyperITE versions to perform better from the first epoch resulting in better estimates.

\end{document}